\pdfoutput=1

\documentclass{article}

     \usepackage[preprint]{neurips_2021}

\usepackage[utf8]{inputenc} 
\usepackage[T1]{fontenc}    
\usepackage{hyperref}       
\usepackage{url}            
\usepackage{booktabs}       
\usepackage{amsfonts}       
\usepackage{nicefrac}       
\usepackage{microtype}      
\usepackage{xcolor}         

\usepackage{amsmath}
\usepackage{amssymb}
\usepackage{shortcuts}
\usepackage[ruled]{algorithm2e} 
\usepackage{algpseudocode}
\usepackage{graphicx}
\usepackage{scalerel}
\usepackage{multirow}
\usepackage{mathtools}

\title{Fast and More Powerful Selective Inference for Sparse High-order Interaction Model}

\author{%
 Diptesh Das \\
 Nagoya Institute of Technology\\
  \texttt{das.diptesh@nitech.ac.jp} \\
 \And
 Vo Nguyen Le Duy\\
 Nagoya Institute of Technology / RIKEN \\
 \texttt{duy.mllab.nit@gmail.com} \\
 \And
 Hiroyuki Hanada \\
 RIKEN \\
 \texttt{hiroyuki.hanada@riken.jp} \\
\And
Koji Tsuda \\
 University of Tokyo / RIKEN \\
 \texttt{tsuda@k.u-tokyo.ac.jp} \\
\And
 Ichiro Takeuchi\\
 Nagoya Institute of Technology / RIKEN  \\
 \texttt{takeuchi.ichiro@nitech.ac.jp} \\
}

\begin{document}

\maketitle


\begin{abstract}
    Automated high-stake decision-making such as medical diagnosis requires models with high interpretability and reliability. As one of the interpretable and reliable models with good prediction ability, we consider Sparse High-order Interaction Model (SHIM) in this study. However, finding statistically significant high-order interactions is challenging due to the intrinsic high dimensionality of the combinatorial effects. Another problem in data-driven modeling is the effect of "cherry-picking" a.k.a. selection bias. Our main contribution is to extend the recently developed parametric programming approach for selective inference to high-order interaction models. Exhaustive search over the cherry tree (all possible interactions) can be daunting and impractical even for a small-sized problem. We introduced an efficient pruning strategy and demonstrated the computational efficiency and statistical power of the proposed method using both synthetic and real data. 

\end{abstract}

\section{Introduction}
Blackbox models such as deep neural network models generally have high predictive performance but are difficult to interpret and hence, often considered unreliable. Therefore, for tasks that require high-stake decision-making, such as medical diagnosis and automated driving, models with higher interpretability and reliability are required. As one of the interpretable and reliable models with good prediction ability, we consider Sparse High-order Interaction Model (SHIM) in this study. Considering a regression problem with a response $y$ and $m$ original covariates $z_1, \ldots, z_m$, an example SHIM up to $4^{th}$ order interactions can be written as 
\begin{equation}\label{eq:shim_eg1}
y = \beta_1 z_3 +  \beta_2 z_5 +  \beta_3 z_2z_6 +  \beta_4 z_1z_2z_5z_9 .
\end{equation}
where \(\beta_1, \beta_2, \beta_3, \beta_4\) are the model parameters (or coefficients). Such a SHIM has practical importance, such as identifying complex genotypic features for HIV-1 drug resistance \citep{saigo2007mining}. HIV-1 evolves in the human body and exposure to certain drugs causes mutations that leads to resistance against the drugs. Structural biological studies show that it is the association of multiple mutations along with some crucial single mutations that can best describe the complex biological phenomenon of drug resistance 
\citep{vivet2006nucleoside, iversen1996multidrug, rhee2006genotypic}.
%

The goal of this study is to fit a SHIM such as (\ref{eq:shim_eg1}) to the given data and subsequently perform statistical significance test to judge the reliability of the model parameters. However, unless the original dimension and the order of interactions are small, fitting a high-order interaction model can be challenging and one would require some computational tricks to avoid the combinatorial effects. 

Another challenge of data-driven modeling is understanding the reliability of findings because the model might have cherry-picked the strong associations given a particular realization of the data.
This is called "cherry-picking" effect a.k.a. selection bias \citep{taylor2015statistical}.
Traditional statistical inference, which assumes that the statistical model and the target for which inferences are conducted must be fixed {\it a priori}, cannot be used for this problem. 
%
Any inference conducted after model selection will suffer from the selection bias unless it is corrected.  
%

%
\textbf{Related works:}
Several approaches have been suggested in the literature to address the above problem ~(\cite{fithian2014optimal, fithian2015selective}, \cite{choi2017selecting}, \cite{tian2018selective}, \cite{chen2020valid}, \cite{hyun2018post}, \cite{loftus2014significance, loftus2015selective}, \cite{panigrahi2016bayesian}, \cite{tibshirani2016exact},  \cite{yang2016selective}).
A particularly notable approach is \emph{conditional} SI introduced in the seminal paper by \citet{lee2016exact}. The basic idea of conditional SI is to make inference on a data-driven hypothesis conditional on the selection event that the hypothesis is selected.
\cite{lee2016exact} first proposed conditional SI methods for the selected features by using Lasso. 
Their basic idea is to characterize the selection event by a polytope, i.e., a set of linear inequalities, in the sample space.
When a selection event can be characterized by a polytope, practical computational methods developed by these authors can be used for making inferences of the selected hypotheses conditional on the selection events. 

However, the conditional SI framework based on a polytope has a serious drawback called \emph{over-conditioning} issue, i.e., additional extra events must be introduced to characterize the selection event by a single polytope, which is known to lead loss of statistical power or \emph{statistically sub-optimal} ~\cite{fithian2014optimal}.
The work by \cite{suzumura2017selective}, who first applied polytope-based SI into high-order interaction model when a high-order interaction feature is sequentially added to the model, also suffers from this problem. As a solution in the case of LASSO \cite{lee2016exact} proposed to take the union of all possible signs of the selected features. However, unless the number of the selected features is small, it is computationally expensive and, in the case of SHIM type problem, it will be impractical due to the combinatorial effects.
%

Recently, \cite{le2021parametric} introduced a homotopy method to resolve the over-conditioning issue and realizes minimally-conditioned SI for Lasso. 
Our basic idea for identifying statistically reliable high-order interaction features in sparse modeling framework is to employ \emph{exact homotopy}-based SI method for SHIM. Unfortunately, the computational cost for applying the exact homotopy method to SHIM increases exponentially and intractable unless the size of the selected features and the maximum order of interactions are fairly small. Several methods have already been proposed for fitting a SHIM \citep{saigo2009gboost, ERP_Tsuda, nakagawa2016safe}.
%


\textbf{Contribution:}
Our main contribution in this paper is to introduce a ``homotopy mining'' method by exploting the best of both homotopy and (pattern) mining methods for conditional SI for SHIM. 
This approach is motivated by the exact regularization path computation algorithm for graph data~\citep{ERP_Tsuda}, which is considered as a homotopy method with respect to the regularization parameter. 
In the algorithm of our proposed method, we use two types of homotopy mining methods, one for fitting a SHIM on the observed dataset (which is essentially the same as the approach in \cite{ERP_Tsuda}) and, another for computing the sampling distribution of the test-statistic conditional on the selection event.
Interestingly, these two types of homotopy mining methods share many common properties such as branch and bound techniques for pruning high-order interaction tree (see Fig.\ref{fig:tree}). We applied our proposed method on synthetic and real-world HIV1 drug resistance data and demonstrated in \S4 that we could quantify the statistical significance of high-order interaction features in the forms of $p$-values and confidence intervals without any computational nor statistical approximations. In an experimental study of the inference stage, we showed that a single traversal of a search space of more than $10^{10}$ high-order interaction terms (sample size, $n=625$) took less than 240 sec (worst case) and 78 sec (best case) on average  using Intel Xeon Gold 6230 CPU @ 2.10 GHZ. We extended this framework to solve the Elastic Net optimization problem which was not trivial as we cannot follow the common approach of data augmentation by stacking extra rows as this can be prohibitively expensive due to the combinatorial effects.
\section{Problem Statement}
%
%
Consider a regression problem with a response vector $y \in \RR^n$ and 
$m$ original covariate vectors $z_1, \ldots, z_m$, where \(z_{j} \in \mb{R}^n\) and $j \in [m] = \{1, ..., m\}$. Then, a high-order interaction model up to $d^{\rm th}$ order is written as
\begin{equation}\label{eq:shim_model}
\begin{split}
\hspace{-2.5mm} y = 
\sum_{j_1 \in [m]} \alpha_{j_1} z_{j_1} 
+ \sum_{ \substack{ (j_1, j_2)  \in [m] \times [m]  \\ j_1 \neq j_2}} \hspace{-2.5mm} \alpha_{j_1, j_2} z_{j_1} \circ z_{j_2} 
+  \cdots + 
\sum_{\substack{(j_1, ..., j_d)  \in [m]^d \\ j_1 \neq ... \neq j_d}} \hspace{-2.5mm} \alpha_{j_1, \ldots, j_d} z_{j_1} \circ \cdots \circ z_{j_d},
\end{split}
\end{equation}
where $\circ$ is the element-wise product and scalar $\alpha$s are the coefficients.
In this paper, we consider each element of the original covariate vector $z_{j} \in \RR^n$, $j \in [m]$, is defined in a domain $[0, 1]$.
To simplify notation, it is convenient to write the high-order interaction model in  (\ref{eq:shim_model}) by using the following matrix of concatenated vectors of all high-order interactions:
\[
X = [\underbrace{z_1, \ldots, z_m}_{1\text{\ts{st} order}}, \underbrace{ z_1 z_2, \ldots, z_{m-1} z_m}_{2\text{\ts{nd}order}}, 
\cdots,
\underbrace{z_1 \ldots z_d, \ldots, z_{m-d+1} \ldots z_m}_{d\text{\ts{th} order}}] \in \RR^{n \times p},
\]
where \(p \coloneqq \sum_{\kappa=1}^d {m \choose \kappa}\).
Similarly, the coefficient vector associated with all possible high-order interaction terms can be written as:
\[
\beta:= [ \underbrace{ \alpha_1, \ldots, \alpha_m}_{1\text{\ts{st} order}},  \underbrace{\alpha_{1,2}, \ldots, \alpha_{m-1, m}}_{2\text{\ts{nd} order}}, 
\cdots,
\underbrace{\alpha_{1, \ldots ,d},  \ldots, \alpha_{m-d+1, \ldots, m}}_{d\text{\ts{th} order}}]^\top \in \RR^p.
\]
%
The high-order interaction model (\ref{eq:shim_model}) is then simply written as a linear model
$
y = X \beta.
$
%
%
Unfortunately, $p$ can be prohibitively large unless both $m$ and $d$ are fairly small.
%
%
%
In SHIM, we consider a sparse estimation of high-order interaction model.
An example of SHIM looks like
\begin{equation}\label{eq:shim_eg}
 y = \alpha_3z_{3} +  \alpha_5z_{5} +  \alpha_{2,6}z_{2}z_{6} +  \alpha_{1,2,5,9}z_{1}z_{2}z_{5}z_{9}.
\end{equation}

The goal of this study is to fit a SHIM such as (\ref{eq:shim_eg}) and test the statistical significance of the coefficients of the selected model (in the above example, \(  \alpha_3,  \alpha_5, \alpha_{2,6}, \alpha_{1,2,5,9} \)) in order to quantify the reliability.
%
%
Unfortunately, both fitting and testing a SHIM are non-trivial because, unless both $m$ and $d$ are very small, a high-order interaction model will have an extremely large number of parameters to be considered.
Several algorithms for fitting a sparse high-order interaction model were proposed in the literature (see \S1).
A common approach taken in these existing works is to exploit the hierarchical structure of high-order interaction features.
In other words, a tree structure as in Fig.~\ref{fig:tree}(a) is considered and a branch-and-bound strategy is employed in order to avoid handling all the exponentially increasing number of high-order interaction features.

Here, we introduce an alogrithm for conditional SI in order to quantify the statistical significance of the fitted coefficients of SHIM such as $\alpha_3,  \alpha_5, \alpha_{2,6}, \alpha_{1,2,5,9}$ in the forms of $p$-values or confidence intervals by using homotopy-based SI.
However, due to the extremely large number of features in \eq{eq:shim_model}, it is intractable to characterize the selection event for homotopy-based SI. 
In order to overcome this challenge, we develop \emph{homotopy mining} method which effectively combines the homotopy method and branch-and-bound strategy in the cherry tree.
Before delving into our proposed method, we briefly overview conditional SI.

\subsection{Selective Inference and Homotopy Method}
\begin{figure}[t]
   \centering
   \includegraphics[width=\linewidth]{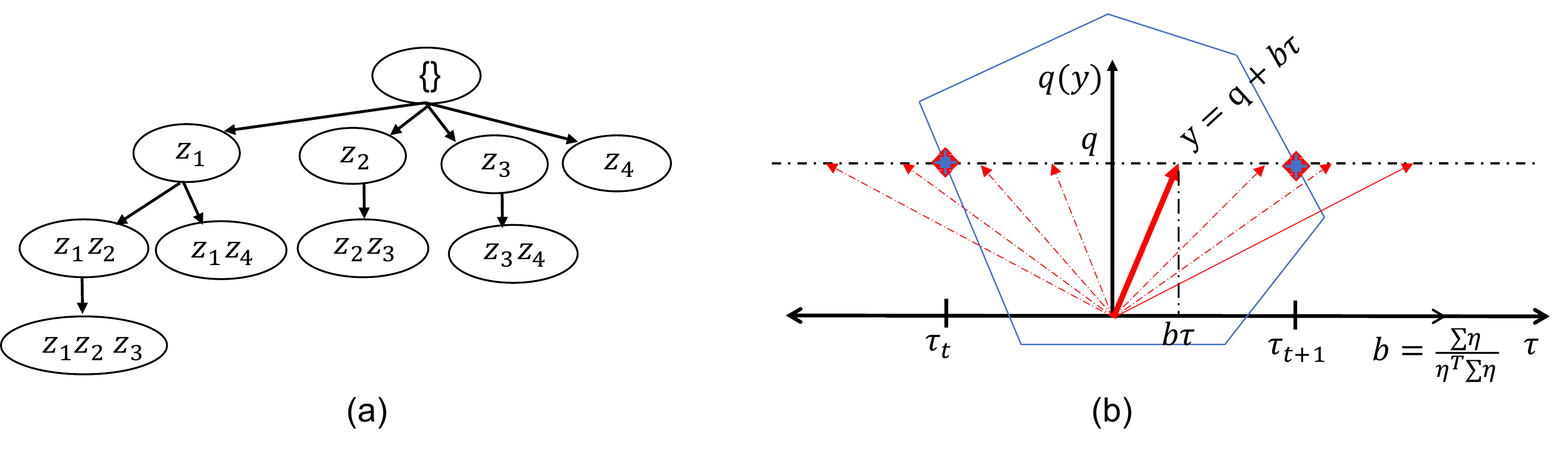}
   \caption{(a) A cherry tree of patterns has been constructed by exploiting the hierarchical structure of high-order interaction features. Not all nodes are traversed due to pruning. (b) The conditional data space is restricted to a line. In the figure it is restricted along the horizontal "$\tau$-line" and we need to find the truncation points ($\tau_t, \tau_{t+1}$) along this line.}
   \label{fig:tree}
 \end{figure}

We present conditional selective inference (SI) which is introduced in \citet{lee2016exact} and then explain that \emph{optimal (i.e., minimally-conditioned)} conditional SI can be conducted with a homotopy method.
%
%
In conditional SI framework, we assume that the design matrix $X$ is fixed, response vector $y$ is a realization of random response vector $Y \sim N(\mu, \Sigma)$, where $\mu \in \RR^{n}$ is unknown mean vector and $\Sigma \in \RR^{n \times n}$ is covariance matrix which is known or estimable from external data.
In this framework, we do not assume ``true'' relationship between $X$ and $\mu$, but consider a case where the data analyst adopts the SHIM as a reasonable approximation model to describe the relationship.

Let $\cA$ be the set of selected features by solving the SHIM fitting problem. 
With a slight abuse of notation, we also write this set of features as $\cA(y)$ in order to emphasize that the set of features $\cA$ is obtained when $y$ is observed. 
This notation enables us to consider $\cA(y^\prime)$ as the set of features which would be selected when a different response vector $y^\prime$ is observed.
Furthermore, $\cA(Y)$ represents the ``random'' set of features selected from the ``random'' response vector $Y$.

Given the set of selected features $\cA$, consider the best linear approximation of $\mu$ with the selected features.
For $j \in \cA$, let
\begin{align*}
\beta_j^* := (X_\cA^\top X_\cA)^{-1} X_{\cA}^\top \mu 
\end{align*}
be the $j^{\rm th}$ population coefficient of the best linear approximation model fitted only with the selected features. 
In conditional SI framework, we consider the following hypothesis test:
\begin{align}
 \label{eq:statistical_test}
 {\rm H}_0: \beta_j^* = 0 ~~~\text{v.s.}~~~ {\rm H}_1: \beta_j^* \neq 0, ~ j \in \cA.
\end{align}
Noting that, by defining $\eta := e_j^\top X_{\cA(Y)} (X_{\cA(Y)}^\top X_{\cA(Y)})^{-1}$ with $e_j \in \RR^n$ being the vector with 1 at the $j^{\rm th}$ component and 0 otherwise, we can write $\beta_j^* = \eta^\top \mu$ with $Y = y$. 
Therefore, it is reasonable to use $\eta^\top Y$ as the test statistic for the test \eq{eq:statistical_test}.
The (unconditional) sampling distribution of $\eta^\top Y$ is highly complicated and intractable because $\eta$ also depends on the random response vector $Y$ through the selected features $\cA(Y)$. 
The basic idea of conditional SI is to consider the sampling distribution of the test-statistic conditional on the selection event, i.e., $\eta^\top Y \mid  \{\cA(Y) = \cA\}$.
%
By further conditioning on the nuisance component $q(Y) = (I_n - b \eta^\top) Y$ with $b := \Sigma \eta (\eta^\top \Sigma \eta)^{-1}$ which is independent of the test statistic $\eta^\top Y$, \cite{lee2016exact} showed that the conditional sampling distribution of $\eta^\top Y \mid \{\cA(Y) = \cA, q(Y) = q \}$ follows a truncated Normal distribution 
\begin{equation}\label{eqn:conditional_sampling_distr}
 \eta^\top Y \mid \{\mathcal{A}(Y)=\mathcal{A}, q(Y)=q\} \sim F^{\mathcal{T}}_{\eta^T\mu,\eta^T\sum\eta},
\end{equation}
where $F_{\tilde{\mu}, \tilde{\sigma}^2}^{\cT}$ is the c.d.f. of the truncated Normal distribution with mean $\tilde{\mu}$, variance $\tilde{\sigma}^2$, the truncation region $\cT$, and $q$ is the observed nuisance component defined as $q = (I_n - b \eta^\top) y$. However, identifying the conditional data space \(\{\mathcal{A}(Y)=\mathcal{A}, q(Y)=q\}\) is a challenging problem.

In \cite{lee2016exact}, the authors developed a practical algorithm to compute the truncated Normal distribution by further conditioning on the signs of the selected features in $\cA$. 
Although the validity of the inference can be maintained with this additional conditioning on the signs, it turns out that the power of the inference is \emph{suboptimal} with this over-conditioning~\citep{fithian2014optimal}. Recently, \citet{le2021parametric} developed an algorithm to resolve this issue by using homotopy method. 
In particular, they considered the parametrized response vector (see Fig. \ref{fig:tree} (b))
\begin{equation}\label{eq:parametrized_response_vector}
  y(\tau) := q + b \tau   
\end{equation}
for a scalar parameter $\tau \in \RR$, and solve the continuum of optimal solutions when the response vector $y$ is replaced with $y(\tau)$ by using homotopy method. Therefore, we can redefine the conditional data space in (\ref{eqn:conditional_sampling_distr}) as
\begin{equation} \label{eq:def_truncation_region}
 \mcl{T} = \{\tau \in \mb{R} \hst | \hst \mcl{A}(y(\tau)) = \mcl{A}(y) \}. 
\end{equation}
It enables us to completely identify the truncation region of the truncated Normal sampling distribution and compute the selective $p$-value 
\begin{equation}
 P_j^{\rm selective} = 2 \hspace{0.1cm} \text{min} \{ \pi_j, 1 - \pi_j\}, \quad \text{where,} \quad \pi_j = 1 - F^{\mathcal{T}}_{0,\eta^T\sum\eta} (\eta^\top y).
\end{equation}
Similarly, one can obtain $1 - \alpha$ confidence interval $\mcl{C}_\alpha$ for any $\alpha \in [0, 1] $ such that 
\begin{equation*}
\mb{P}(\beta_j^* \in \mcl{C}_\alpha \big| \{\mathcal{A}(Y)=\mathcal{A}, q(Y)=q \}) = 1 -\alpha. 
\end{equation*}
Unfortunately, in the case of SHIM, since the number of high-order interaction features are exponentially large, we cannot use the same homotopy method. In the following section, we present the \emph{homotopy mining algorithm} which enables us to compute the conditional sampling distribution (\ref{eqn:conditional_sampling_distr}) of the fitted SHIM coefficients by effectively combining homotopy method and branch-and-bound method in pattern mining.

 \section{Proposed Method}\label{homotopy-mining} 
 In this study we propose a similar \textit{``homotopy-mining''} approach for model selection and inference. Homotopy method refers to an optimization framework for solving a sequence of parameterized optimization problems. The basic idea of our homotopy mining approach is to consider the following optimization problem with a parameterized response vector $y (\tau)$ in (\ref{eq:parametrized_response_vector})
 %
\begin{equation}\label{eq:shim_homotopy_opt}
\beta(\lambda, \tau) = \argmin_{\beta \in \bbR^p} \hspace{0.2cm} \mathcal{F}_{\lambda, \tau}(\beta) :=  \frac{1}{2}\norm{y(\tau) - X\beta}^2 + \lambda \norm{\beta}_1,
\end{equation}
where $\tau \in \mathbb{R}$ is a scalar parameter, $\lambda$ is the regularization parameter for $L_1$-regularization, and the objective function $\mathcal{F}_{\lambda, \tau}(\beta)$ is parameterized by both $\tau$ and $\lambda$. 
The homotopy mining enables us to solve a sequence of parameterized optimization problems in the form of (\ref{eq:shim_homotopy_opt}) by effectively combining homotopy and mining method. 

To extend the homotopy selective inference framework for SHIM, we first need to solve (\ref{eq:shim_homotopy_opt}) for a fixed $\tau$ and target $\lambda$ using the observed data and obtain an active set $\mathcal{A}$. 
Now, $\forall j  \in \mathcal{A}$, we need to construct the exact solution path characterized by $\tau$ and then identify the conditional data space in (\ref{eq:def_truncation_region}) by identifying the intervals of $\tau$ on the solution path. 
This exact solution path can be constructed in a similar manner as the LARS-LASSO algorithm by an efficient step size calculation. 
 Here, we define the exact regularization paths \(\lambda \mapsto \beta(\lambda)\) for a fixed $\tau$ as the ``\(\lambda\)-\textit{path}'' and \(\tau \mapsto \beta(\tau)\) for a fixed $\lambda$ as the ``\(\tau\)-\textit{path}'', respectively. 
Then, both the selection and inference paths of the SHIM can be constructed in a similar fashion as stated below:

$\bullet$ Model selection of SHIM can be done by using exact regularization path algorithm 
 \begin{equation}\label{eq:lambda_seq_path}
 \lambda_0 >  \lambda_1 > \cdots  > \lambda_{\rm min}  \Rightarrow  \{\beta(\lambda_0), \beta(\lambda_1) , \cdots, \beta(\lambda_{\rm min}) \}.
  \end{equation}
 $\bullet$ For inference, we can have similar path algorithm 
  \begin{equation}\label{eq:z_seq_path}
  \tau_0 >  \tau_1 >  \cdots  > \tau_{\rm min }  \Rightarrow  \{\beta(\tau_0), \beta(\tau_1), \cdots, \beta(\tau_{\rm min}) \},
  \end{equation}
where sequences of $\lambda$ and $\tau$ represent the breakpoints of homotopy method.
The Equations (\ref{eq:lambda_seq_path}) and (\ref{eq:z_seq_path}) have similar problem structure, the only difference is that in (\ref{eq:lambda_seq_path}) we find the solution path characterized by the regularization parameter $\lambda$, whereas in (\ref{eq:z_seq_path}) we find the solution path characterized by $\tau$. 
 Basically, what we need to characterize the selection event is to find those breakpoints (e.g. $\tau_0, \tau_3, \tau_8$) along  the $\tau$-line where the active set remains the same as the observed one, i.e.,
$
 \mathcal{A}(y) =  \mathcal{A}(y(\tau_0)) = \mathcal{A}(y(\tau_3)) =  \mathcal{A}(y(\tau_8)).
  $
  %
 %
 However, computing the exact regularization paths for such SHIM is a challenging task due to exponentially expanded feature space. 
 Efficient computational methods are required both at the selection and inference stage. 
 %
Therefore, we considered a tree structure (see Fig. \ref{fig:tree} (a)) of the interaction terms (or patterns) and proposed a tree pruning strategy both for the selection path ($\lambda$-path) and inference path ($\tau$-path). 
In the next section, we will present the main technical details of  characterizing the conditional data space in (\ref{eq:def_truncation_region}) by using homotopy-mining method.
 
\subsection{Characterization of truncation region in SHIM}
The optimal condition of (\ref{eq:shim_homotopy_opt}) can be written as
\begin{equation}\label{eq:opt_condn_shim_homotopy}
    X^\top \big(X\beta(\lambda,\tau) - y(\tau)\big) + \lambda s(\lambda,\tau) = 0 \text{ where} \hst  s_{j}(\lambda, \tau) \in \begin{cases}
        \{ -1, +1 \} \hst \hst    \text{if} \hst \beta_j(\lambda, \tau) \neq 0,\\
        [-1, +1]  \hst \hst \hst \text{if} \hst   \beta_j(\lambda, \tau) = 0,
\end{cases}
\end{equation}
where $j \in [p]$.
Let us define the active set of features as
\(\label{eq:active_set}
\mcl{A}(y(\tau)) = \left\{j \in [p]:\; \beta_j(\lambda, \tau) \neq 0 \right\}
\). 





\paragraph{The $\tau$-path ($\lambda$ fixed).}
Since $\lambda$ is fixed we drop it from the notation. Now consider two real values $\tau_t$ and $\tau_{t+1}$ ($\tau_{t+1}>\tau_t$) at which the active set does not change and their signs also remain the same.
For notational simplicity, we denote $\cA_{\tau_t} = \cA(y(\tau_t))$.
Then, one can write from (\ref{eq:opt_condn_shim_homotopy})
\begin{align}
  \beta_{\mcl{A}_{\tau_t}} (\tau_{t + 1}) - \beta_{\mcl{A}_{\tau_t}} (\tau_t) &= \nu_{\mcl{A}_{\tau_t}} (\tau_t) \times (\tau_{t+1} - \tau_t) \label{eqn:beta_piece_wise_constant} \\ 
  \lambda s_{\mcl{A}^c_{\tau_t}} (\tau_{t + 1}) - \lambda s_{\mcl{A}^c_{\tau_t}} (\tau_t) &= \gamma_{\mcl{A}^c_{\tau_t}} (\tau_t) \times (\tau_{t+1} - \tau_t) \label{eqn:gamma_piece_wise_constant}
\end{align}
%
%
%
%
%
where \(\nu_{\mcl{A}_{\tau_t}} (\tau) = (X_{\mcl{A}_{\tau_t}}^\top X_{\mcl{A}_{\tau_t}})^{-1}X_{\mcl{A}_{\tau_t}}^\top b \) and \(\gamma_{\mcl{A}^c_{\tau_t}} (\tau) = X_{\mcl{A}^c_{\tau_t}}^\top b -  X_{\mcl{A}^c_{\tau_t}}^\top X_{\mcl{A}_{\tau_t}}\nu_{\mcl{A}_{\tau_t}} (\tau) \)  remain constant for all real values of \(\tau \in [\tau_t, \tau_{t+1})\). 
Thus, Equations (\ref{eqn:beta_piece_wise_constant}) and (\ref{eqn:gamma_piece_wise_constant}) state that \(\beta(\tau)\) and \(\lambda s(\tau)\) are piecewise linear in \(\tau\) for a fixed \(\lambda\). 
The derivations of  \(\nu_{\mcl{A}_{\tau_t}} (\tau_t) \) and \(\gamma_{\mcl{A}^c_{\tau_t}} (\tau_t) \) are given in Appendix A. 
If $\tau_{t+1} > \tau_t$ is the next zero crossing point, then either of the following two events happens

$\bullet$ A zero variable becomes non-zero, i.e., \( \exists j \in \mcl{A}^c_{\tau_t} \text{ s.t. } |x_{j}^{\top}(y(\tau_{t+1}) - X_{\cA_{\tau_{t}}}\beta_{\cA_{\tau_t}}(\tau_{t + 1}))| = \lambda \hst \text{or,}\)

$\bullet$ A non-zero variable becomes zero, i.e.,
\(\exists j \in \mcl{A}_{\tau_t} \text{ s.t. } \beta_j(\tau_t) \neq 0 \text{ and } \beta_j(\tau_{t+1}) = 0 \enspace.\)

Overall, the next change of the active set happens at $\tau_{t + 1} = \tau_t + \Delta_j$, where 
\begin{equation}\label{eq:step_length}
\Delta_j = \min(\Delta_j^1, \Delta_j^2) = \min\left(\min_{j \in \mathcal{A}^c_{\tau_t}} \Big( \lambda \frac{ \text{sign}(\gamma_j (\tau_t)) - s_j(\tau_t)}{\gamma_j(\tau_t)} \Big)_{++}, \hst \min_{j \in \mcl{A}_{\tau_t}} \Big( - \frac{\beta_j(\tau_t)}{\nu_j(\tau_t)} \Big)_{++} \right) \enspace.
\end{equation}
Here, we use the convention that for any $a \in \RR$, $(a)_{++} = a$ if $a > 0$ and $\infty$ otherwise.
The derivation of the step-size $\Delta_j$ for the $\tau$-path is given in the Appendix A. However, solving the minimization problem to determine the step-size of the $\tau$-path and the $\lambda$-path (the details of $\lambda$-path are given in Appendix A) can be challenging for SHIM type problems. 
Hence, we need efficient computational methods to make it practically feasible. 
In the following section we present an efficient tree pruning strategy by considering a tree structure of the interaction terms (or patterns). 
Similar pruning strategy already exists in the literature to solve the $\lambda$-path of the LASSO in the context graph mining [\cite{ERP_Tsuda}]. In the next section we will show that the same pruning strategy can be applied for the $\tau$-path of the SHIM.

\subsection{Tree pruning}

A tree is constructed in such a way that for any pair of nodes (\(\ell, \ell^\prime)\), where $\ell$ is the ancestor of $\ell^\prime$, i.e., \(\ell \subseteq \ell^\prime\), the following conditions are satisfied 
 \begin{equation*}
    \hst  x_{i \ell^\prime} = 1 \implies x_{i\ell } = 1  \hst
 \text{ and conversely,} \hst
     x_{i\ell } = 0 \implies x_{i \ell^\prime} = 0  \quad \forall i \in [n].
 \end{equation*}
Now considering the $\tau$-path of the LASSO, the equicorrelation condition for any active feature \(k  \in \mathcal{A}_{\tau_{t + 1}}\) at a fixed $\lambda$ can be written as
\begin{equation*}
    \left|x_k^\top (y(\tau_{t + 1}) - X\beta(\tau_{t + 1}))\right| = \lambda.
\end{equation*}
Therefore at a fixed \(\lambda\), any non-active feature \(\ell \in \mathcal{A}^c_{\tau_{t} } \) becomes active at $\tau_{t + 1}$ when the following condition is satisfied
\begin{align}\label{eqn:inclusion_condition2}
    \big\lvert x_\ell^\top \big(y(\tau_{t + 1}) - X_{\mathcal{A}_{\tau_t}}(\beta_{\mathcal{A}_{\tau_t}} (\tau_t) + \Delta_\ell \nu_{\mathcal{A}_{\tau_t}} (\tau_t)) \big)\big\rvert 
    &= 
    \big\lvert x_k^\top \big(y(\tau_{t + 1}) - X_{\mathcal{A}_{\tau_t}}(\beta_{\mathcal{A}_{\tau_t}} (\tau_t) + \Delta_\ell \nu_{\mathcal{A}_{\tau_t}} (\tau_t)) \big)\big\rvert \nonumber \\
    \text{or} \quad |\rho_\ell(\tau_t, \tau_{t + 1}) - \Delta_\ell \eta_\ell(\tau_t) | &= |\rho_k(\tau_t, \tau_{t + 1}) - \Delta_\ell \eta_k(\tau_t)|,
\end{align}
where the l.h.s. corresponds to \(\ell \in \mathcal{A}_{\tau_t}^c \) and the r.h.s. corresponds to \(k \in \mathcal{A}_{\tau_t} \). 
%
%
Here, we define
\(\rho_\ell (\tau_{t}, \tau_{t + 1}) = x_\ell^\top \Big(y(\tau_{t + 1}) - X_{\mathcal{A}_{\tau_t}} \beta_{\mathcal{A}_{\tau_t}} (\tau_t) \Big)  \text{ and } \eta_\ell(\tau_t) = x_\ell^\top X_{\mathcal{A}_{\tau_t}} \nu_{\mathcal{A}_{\tau_t}}(\tau_t) \). 
The r.h.s. of (\ref{eqn:inclusion_condition2}) has a lower bound, i.e.,
\[
    |\rho_k(\tau_t, \tau_{t + 1}) - \Delta_\ell \eta_k(\tau_t)| \geq |\rho_k(\tau_t, \tau_{t+ 1})| - \Delta_\ell |\eta_k(\tau_t)|,
\] and the l.h.s. of (\ref{eqn:inclusion_condition2}) has an upper bound, i.e., 
\[
    |\rho_\ell(\tau_{t}, \tau_{t + 1}) - \Delta_\ell \eta_\ell(\tau_t) | \leq |\rho_\ell(\tau_t, \tau_{t+1})| + \Delta_\ell |\eta_\ell(\tau_t)|.
\]
Therefore, for equation (\ref{eqn:inclusion_condition2}) to have a solution, the following condition needs to be satisfied
\begin{equation}\label{eqn:solution_condn}
 \quad \quad  |\rho_\ell(\tau_t, \tau_{t + 1})| + \Delta_\ell |\eta_\ell(\tau_t)| \geq  |\rho_k(\tau_t, \tau_{t + 1})| - \Delta_\ell |\eta_k(\tau_t)|.
\end{equation}
If the above condition (\ref{eqn:solution_condn}) is not satisfied, then equation (\ref{eqn:inclusion_condition2}) will not have any solution, and that can be used as a pruning condition. Therefore, the pruning condition can be written as
\begin{equation}\label{eqn:pruning_condn}
    |\rho_\ell(\tau_t, \tau_{t + 1})| + \Delta_\ell |\eta_\ell(\tau_t)| <  |\rho_k(\tau_t, \tau_{t + 1})| - \Delta_\ell |\eta_k(\tau_t)|.
\end{equation}



\begin{lemma}\label{lemma:lemma_1}
If $\Delta^\ast_\ell$ is the current minimum step-size, i.e. \(\Delta^\ast_\ell = \underset{t \in \{1, 2, \ldots, \ell\}}{\min}\{\Delta_t\}, \)
(\ref{eqn:pruning_condn}) is equivalent to  
\begin{equation*}
    |\rho_{\ell}(\tau_t, \tau_{t + 1})| + \Delta^\ast_\ell |\eta_\ell(\tau_t)| <  |\rho_k(\tau_t, \tau_{t + 1})| - \Delta^\ast_\ell |\eta_k(\tau_t)|.
\end{equation*}
\end{lemma}

\begin{lemma}\label{lemma:lemma_2}
If Lemma \ref{lemma:lemma_1} holds, then $\forall \ell^\prime \supset \ell $, 
\begin{equation} \label{eq:pruning_2}
    |\rho_{\ell^\prime}(\tau_t, \tau_{t + 1})| + \Delta^\ast_\ell |\eta_{\ell^\prime}(\tau_t)| <  |\rho_k(\tau_t, \tau_{t + 1})| - \Delta^\ast_\ell |\eta_k(\tau_t)|.
\end{equation}
\end{lemma}

If the Lemma \ref{lemma:lemma_2} holds, then $\forall \ell^\prime \supset \ell $, $\Delta_{\ell^\prime} > \Delta^\ast_{\ell }$.
Therefore, we can use Lemma \ref{lemma:lemma_2} as the pruning criterion to prune the sub-tree with $\ell^\prime$ as the root node.
The proofs of Lemmas \ref{lemma:lemma_1} and \ref{lemma:lemma_2} are deferred to Appendix A.
The complete algorithm for the inference path ($\tau$-path) is given in Algorithm \ref{algo:inference_path}. 


\begin{algorithm}[h!]
\caption{$\tau$-path}
\label{algo:inference_path}
\begin{algorithmic}[1]
\State \textbf{Input:} \(Z, \lambda, b, q, [\tau_{\rm min}, \tau_{\rm max}]\)
\State Initialization: \( t=0, \tau_t=\tau_{\rm min}, \mathcal{T}= \{ \tau_t \} \), \(\beta(\tau_t)=0\)

\State \(y(\tau_t) = q + b \tau_t \), \( \quad \mathcal{A}_{\tau_t}, \beta_{\mathcal{A}_{\tau_k}} (\tau_t) \leftarrow \lambda\text{-path}(Z, y(\tau_k), \lambda)\) (The algorithm of $\lambda$-path is in Appendix A)

\State  \(\nu_{\mc{A}_{\tau_{t}}} (\tau_t) = (X_{\mc{A}_{\tau_{t}}}^\top X_{\mc{A}_{\tau_{t}}})^{-1}X_{\mc{A}_{\tau_{t}}}^\top b\), \quad \(\nu_{\mc{A}_{\tau_{t}}^c}(\tau_t) = 0\)



\State \textbf{while} \((\tau_t < \tau_{max})\) \textbf{do}
    
%
    
    \State Compute step-length $\Delta_j \leftarrow$ Equation (\ref{eq:step_length})
    
    \State If $\Delta_j = \Delta_j^1$, add  \(j\) into \(\mathcal{A}_{\tau_t}\) \Comment{Inclusion}
    
    \State If \(\Delta_j = \Delta_j^2\), remove $j$ from \(\mathcal{A}_{\tau_t}\) \Comment{Deletion}

    \State update: \( \tau_{t+1} \leftarrow \tau_{t} + \Delta_j \), \(\mathcal{T}=\mathcal{T} \cup \{\tau_{t+1}\} \), \( \beta_{\mc{A}_{\tau_{t+1}}} (\tau_t) \leftarrow \beta_{\mc{A}_{\tau_t}} (\tau_t) + \Delta_j \nu_{\mc{A}_{\tau_t}} (\tau_t) \), \(y(\tau_{t+1}) = q + b \tau_{t+1} \),
    \(\nu_{\mc{A}_{\tau_{t+1}}} (\tau_{t + 1}) = (X_{\mc{A}_{\tau_{t+1}}}^\top X_{\mc{A}_{\tau_{t+1}}})^{-1}X_{\mc{A}_{\tau_{t+1}}}^\top b\), \quad \(\nu_{\mc{A}_{\tau_{t+1}}^c} (\tau_{t + 1}) = 0\)
    
    \State \textbf{end while}
    \State \textbf{Output:} \( \mathcal{T}, \{\mathcal{A}_{\tau_t}\}_{\tau_t \in \mathcal{T}} \)
\end{algorithmic}
\end{algorithm}

\subsection{Extension for Elastic Net}
We extended our proposed method to solve the elastic net optimization problem. However, we could not follow the general approach of solving the elastic net optimization problem as solving LASSO with augmented data. Because, we cannot just simply augment the data by stacking extra rows as this can be prohibitively expensive due to the combinatorial effects. 
In order to derive the step-size for both $\lambda$-path and $\tau$-path, we need a different approach as we construct the high-order interaction model in a progressive manner. 
We have shown that using a simple trick, the step-size can be computed very efficiently.
Similar trick is also used to derive the pruning condition. See Appendix B for the details.

\section{Experiments}

We only highlight the main results. The details of experimental setup and several additional experimental results are deferred to Appendix C.

 
 

\subsection{Comparison of statistical powers.}
\textbf{Synthetic data:}
\begin{figure}[t]
   \centering
   \includegraphics[width=\linewidth]{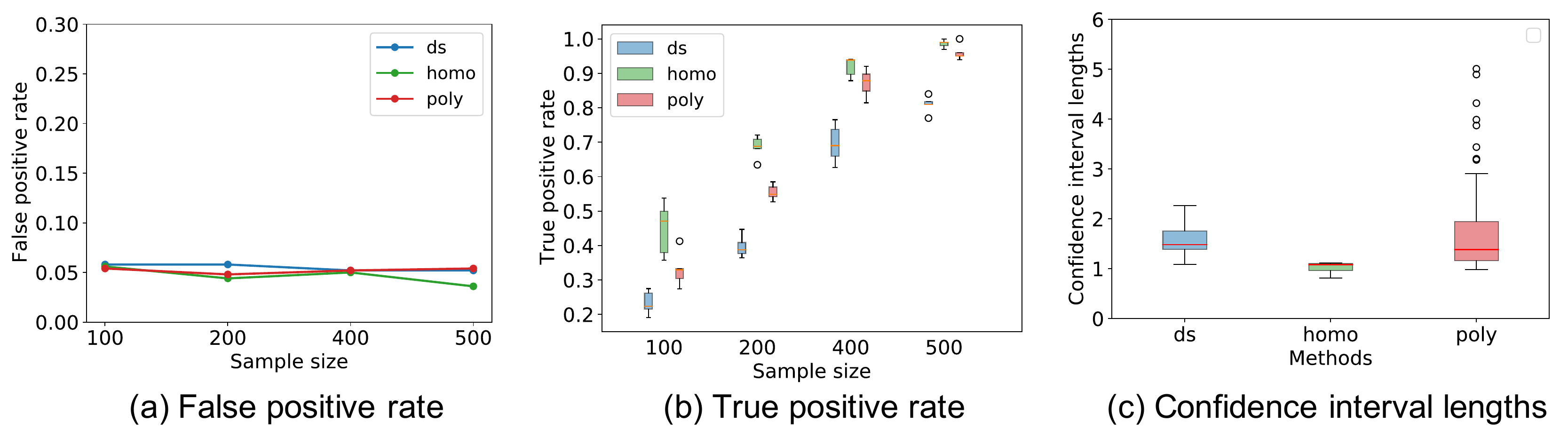}
   \caption{Demonstration of the statistical power of three selection bias correction methods (ds: data splitting, homo: homotopy, poly: polytope) using synthetic data experiments. (a) and (b) show the false positive rates and the true positive rates for different sample sizes and (c) shows the distribution of the confidence interval lengths.}
   \label{fig:stat_power_synthetic}
 \end{figure}
We generated the i.i.d. random samples $(z_i, y) \in [0,1]^m \times \mb{R} $ in such a way that $100m(1-\zeta)\%$ of $z_i \in \mb{R}^m$ contain $1s$on average.
Here, $\zeta \in [0, 1]$ is the sparsity controlling parameter.
The response $y_i \in \mb{R}$ is randomly generated from a normal distribution $N(0, \sigma^2)$.
For the comparison of false positive rates (FPRs), true positive rates (TPRs) and confidence interval (CI) across different methods, we generated the design matrix for a fixed sparsity parameter $\zeta = 0.95$.
In all experiments, the significance level was set as $\alpha = 0.05$.
For the comparison of TPRs we considered a true model of up to $3^{\rm rd}$-order interactions defined as $\mu (x_i) = 0.5z_1 - 2z_2z_3 + 3z_4z_5z_6$.
The response $y_i$ is accordingly generated from $N(\mu(X), \sigma^2I)$.
For the comparison of FPRs, we set $\beta_j = 0, \forall j \in \mb{R}^p$. 
We compared both FPRs and TPRs across three different methods ({\tt ds}: data splitting, {\tt homo}: homotopy, {\tt poly}: polytope) for four different sample sizes $n \in [100, 200, 400, 500]$.
We generated TPRs and FPRs over 100 trials for all three methods and repeated the experiments for $5$ times.
The results are shown in Fig.~\ref{fig:stat_power_synthetic}(a) and Fig.~\ref{fig:stat_power_synthetic}(b), respectively.
It can be seen that all SI methods can properly control the FPRs under $\alpha = 0.05$.
Regarding the TPRs comparison, it can be seen that homotopy has the highest power which is obvious as it is minimally conditioned compared to polytope which suffers from over conditioning.
Comparing TPRs of data splitting ({\tt ds}) and homotopy ({\tt homo}), it can be seen that TPRs of {\tt homo} is always greater than that of {\tt ds}.
Note that in {\tt ds}, only half of the data is used for selection and the remaining half is used for the inference.
Therefore, compared to {\tt homo}, {\tt ds} has higher risk of failing to identify truly correlated features in selection stage and similarly suffer from low statistical power in the inference stage.   
The result of CIs is shown in Fig.~\ref{fig:stat_power_synthetic}(c).
Here, we used the same true model of the TPR experiments and reported the average CIs over 100 trials across different methods.
The results of CIs are consistent with the findings of TPRs.

\begin{figure}[t]
   \centering
   \includegraphics[width=0.95\linewidth]{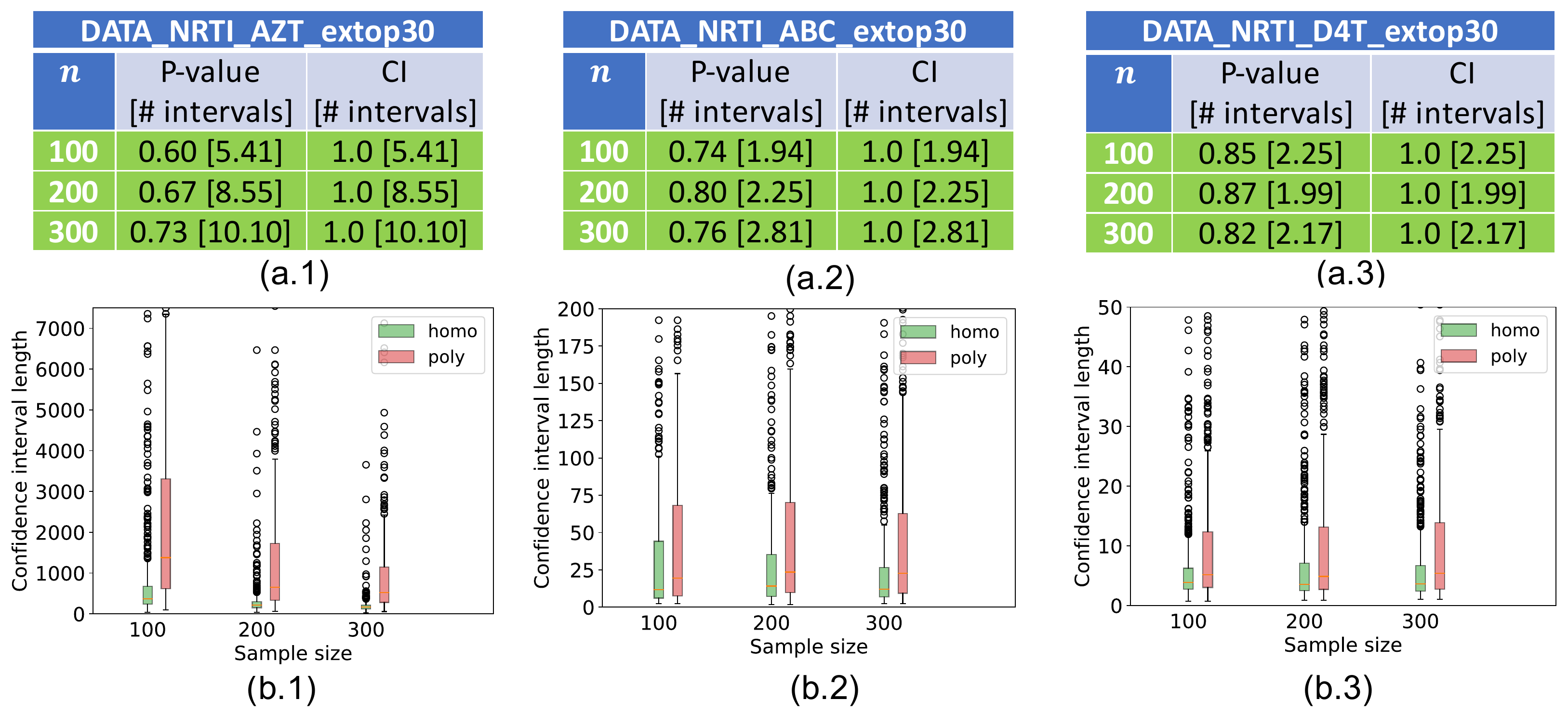}
   \caption{Comparison of statistical powers (Homotopy vs Polytope). (a.1-a.3) show the percentage of cases where selection bias corrected p-values and confidence interval lengths of the proposed method (Homotopy) was smaller than that of the existing method (Polytope) in random sub-sampling experiments. (b.1-b.3) show the distributions of the confidence interval lengths of the same experiments. The numbers inside the brackets represent the average number of intervals along the $\tau$-line considered for the homotopy method. Note that in case of polytope only one such interval is considered.}
   \label{fig:stats_NRTIs1}
\end{figure}
 \textbf{Real data:}
%
%
We obtained HIV-1 sequence data from Stanford HIV Drug Resistance Database \cite{rhee2003human}.
In our experiment we used 6 NRTIs, 1 NNRTIs and  3 PIs drugs.
We only reported here the results of 3 NRTIs drugs.
Additional results are included in the Appendix C.
To demonstrate the statistical efficacy of the proposed homotopy method over existing polytope method we generated random sub-samples of those 10 drug data as follows.
First, we created a dataset consisting of top 30 mutations from each of the 10 drug data.
As most of the columns contain zeros we sorted the columns based on the number of 1's present in each column and picked the top 30 columns as our starting set.
Then, from this starting set we considered random sub-samples of five features for three different sample sizes ($n \in \{100, 200, 300\}$).
Here, we considered randomization without replacement for both sample and features selection.
We generated 100 samples and repeated the experiments for five times and hence, in total we generated 500 samples.
Figure~\ref{fig:stats_NRTIs1} demonstrates the percentage of times homotopy produced smaller $p$-values and CI lengths than the polytope.
This also depicts the distributional difference of the CI lengths between homotopy and polytope.
These results clearly demonstrate that homotopy is statistically more powerful than existing polytope method.
\begin{figure}[t]
   \centering
   \includegraphics[width=\linewidth]{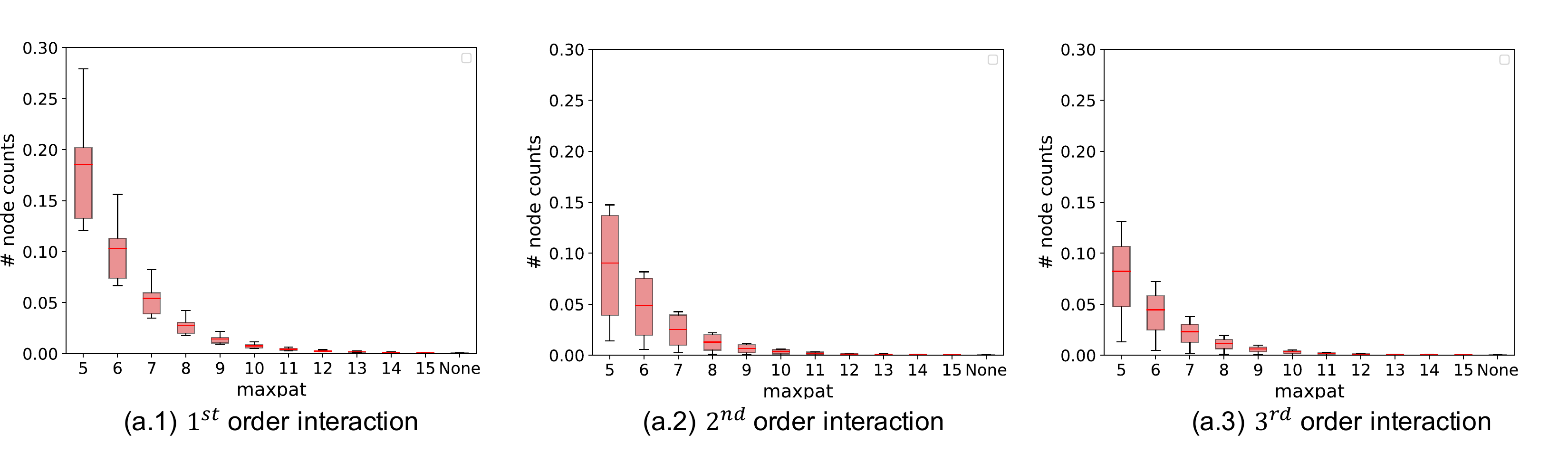}
   \caption{Distribution of the fraction of total nodes traversed against different maximum pattern size ($d$) constraints while applying the proposed pruning method during the construction of the $\tau$-path. (a.1) - (a.3) demonstrate the results for $1^{st}$, $2^{nd}$ and $3^{rd}$ order interaction terms. }
   \label{fig:comp_eff_node_counts}
\end{figure}
\begin{table}[t]
\centering
\resizebox{\textwidth}{!}{
\begin{tabular}{ |c|c|c|c|c|c|c|c| } 
\hline
 \multirow{2}{*}{$d$} & \multirow{2}{*}{\shortstack{Search space \\ (\# nodes)}} &  \multicolumn{3}{|c|}{With pruning} & \multicolumn{3}{|c|}{Without pruning}\\\cline{3-8}
  & & $1^{st}$ & $2^{nd}$ & $3^{rd}$ & $1^{st}$ & $2^{nd}$ & $3^{rd}$ \\
 \hline
  5 & 174436 & $14.56 \pm 6.05$  & $6.58 \pm 2.05$ & $6.78  \pm 3.92 $                 &$25.29 \pm 2.50$  & $34.80 \pm 1.19$  & $23.34 \pm 1.88$ \\
 \hline
 6 & 768211 & $ 34.80 \pm 16.10$  & $13.56 \pm 5.75$  & $13.99 \pm 10.08 $               &$126.96 \pm 8.61$ & $125.29 \pm 2.14$  &  $127.97 \pm 4.80$ \\
 \hline
 7 & 2804011 & $ 68.19 \pm 33.17$  & $24.15 \pm 11.83$  & $25.19  \pm 20.50$             & $450.24 \pm 28.50$ & $447.59 \pm 22.15$ & $447.19 \pm 37.69$  \\
 \hline
 8 & 8656936 & $ 110.25 \pm 55.55$  & $37.70 \pm  19.39$ & $39.45 \pm 33.37$             & > 1 day & > 1 day & > 1 day \\
 \hline
  9 & 8656936 & $ 151.31 \pm 76.81$  & $51.08 \pm 27.09 $ & $54.06 \pm 47.34$            & > 1 day & > 1 day & > 1 day \\
 \hline
 10 & 53009101 & $ 188.26 \pm 95.91$  & $63.66 \pm  34.71$ & $65.49 \pm 58.42$                    & > 1 day & > 1 day & > 1 day \\
 \hline
 
  11 & 107636401 & $ 212.34 \pm 105.54$ & $69.26 \pm 38.49 $ & $74.54 \pm 66.42$                  & > 1 day & > 1 day & > 1 day  \\
 \hline
12 & 194129626 & $ 226.98 \pm 115.71$ & $74.36 \pm  41.20$ & $78.97 \pm  70.33$                        & > 1 day & > 1 day & > 1 day  \\
 \hline
 13 & 313889476 & $ 233.88 \pm 117.25$ & $76.86 \pm  43.10$ & $83.09 \pm 75.13$                        & > 1 day & > 1 day & > 1 day  \\
 \hline
 14 & 459312151 & $ 240.36 \pm 124.79$ & $78.127 \pm 43.44$ & $82.98 \pm 74.13$                         & > 1 day & > 1 day & > 1 day  \\
 \hline
 15 & 614429671 & $ 238.0 \pm  $ 120.35 & $79.67 \pm 44.72 $ & $83.31 \pm 75.16$                       & > 1 day & > 1 day & > 1 day  \\
 \hline
 None & 1073741823 & $ 240.17 \pm 119.76 $ & $78.08 \pm 43.62$ & $82.98 \pm 74.33$             & > 1 day & > 1 day & > 1 day  \\
 \hline
\end{tabular}}
\caption{Computation time (in sec) with and without puning for $1^{\rm st}$, $2^{\rm nd}$ and $3^{\rm rd}$ order interactions. Here, the computation time is measured against different maximum pattern size ($d$) constraints. The last row corresponds to the case when "$d$" is not specified and the whole search space is used for exploration. All computation times are measured on Intel Xeon Gold 6230 CPU @ 2.10GHz.}
\label{table:comp_eff_time_taken}
\end{table}
\subsection{Comparison of computational efficiencies.}
To demonstrate the computational efficiency of the proposed pruning strategy for the $\tau$-path, we applied our homotopy method with and without pruning on HIV NRTI D4T drug resistance data with the same starting set of top 30 mutations as used to demonstrate the statistical power.
Although we varied the $d$ from 5 to $m$, high-order interaction terms upto $3^{\rm rd}$ order appeared in $\mcl{A}$.
We compared both the number of nodes traversed (Fig.\ref{fig:comp_eff_node_counts}) and the time taken (Table.\ref{table:comp_eff_time_taken}) against different maximum interaction order $d$ during the construction of the $\tau$-path of each test statistic direction.
Empirically it was found that the pruning was more effective for the $\tau$-path of high-order interaction terms compared to that of singleton terms and the power of pruning increases as the order of interaction increases. 

Therefore, we reported the average number of nodes and average time taken separately for $1^{\rm st}$, $2^{\rm nd}$ and $3^{\rm rd}$ order interaction terms.
It can be observed that the pruning is more effective at the deeper nodes of the tree and it saturates after certain depth of the tree.
This is evident as the sparsity of the data increases at the deeper nodes and the pruning exploits the monotonicity of high-order interaction terms constructed as tree.
In case of homotopy method without pruning we stopped the execution of program if the $\tau$-path was not finished in one day.
From Tab.~\ref{table:comp_eff_time_taken}, it can be observed that without the pruning the construction of $\tau$-path is not practical owing to the generation of exponential number of high-order interaction terms as we progress to the deeper nodes of the tree.
The $\tau$-path without pruning took more than a day beyond $d=7$, while the maximum time taken by the $\tau$-path with pruning was around 240 sec on average, even when no $d$ constraint was imposed.
\section{Conclusions}
In this paper, we presented an algorithm for testing a sparse high-order interaction model (SHIM) by using the framework of conditional selective inference (SI). The algorithm is developed by effectively combining the homotopy and branch-and-bound tree mining method to deal with the combinatorial computational burden of the SHIM and also to improve the statistical power.

\bibliographystyle{ACM-Reference-Format}
\bibliography{ms}

\appendix
\section{Appendix}
\subsection{LASSO $\tau$-path.}
\subsubsection{Derivations of $\nu_{\cA_{\tau_t}} (\tau_t)$ and $\gamma_{\cA^c_{\tau_t}} (\tau_t)$ in Equations (\ref{eqn:beta_piece_wise_constant}) and (\ref{eqn:gamma_piece_wise_constant}).}\label{LASSO:direction_vector_tau_path}
From the optimality conditions (\ref{eq:opt_condn_shim_homotopy}) of the Lasso at $\tau_t$ and $\tau_{t+1}$, we have following equations for the active components
%
%
%
\begin{align}
	-X^{\top}_{\mathcal{A}_{\tau_t}}(y(\tau_t) - X_{\mathcal{A}_{\tau_t}}\beta_{\mathcal{A}_{\tau_t}}(\tau_t)) 
    + \lambda s_{\mathcal{A}_{\tau_t}}(\tau_{t}) = 0, \label{eqn:lasso_normal_tau1} \\ 
    -X^{\top}_{\mathcal{A}_{\tau_{t}}}(y(\tau_{t+1}) - X_{\mathcal{A}_{\tau_{t}}}\beta_{\mathcal{A}_{\tau_{t}}}(\tau_{t+1})) 
    + \lambda s_{\mathcal{A}_{\tau_{t}}}(\tau_{t}) = 0.\label{eqn:lasso_normal_tau2}
\end{align}
Note that \(\mcl{A}_{\tau_t} = \mcl{A}_{\tau_{t+1}}\) and \(s_{\mc{A}_{\tau_t}}(\tau_{t}) = s_{\mcl{A}_{\tau_{t}}}(\tau_{t+1})\). Therefore, subtracting (\ref{eqn:lasso_normal_tau1}) from (\ref{eqn:lasso_normal_tau2}) we can write
\begin{align*}
    \beta_{\mcl{A}_{\tau_{t}}}(\tau_{t+1}) - \beta_{\mathcal{A}_{\tau_t}}(\tau_t) 
    &= (X^{\top}_{\mc{A}_{\tau_t}}X_{\mc{A}_{\tau_t}})^{-1}X^{\top}_{\mc{A}_{\tau_t}}(y(\tau_{t+1}) - y(\tau_t))\\
    &= (X^{\top}_{\mc{A}_{\tau_t}}X_{\mc{A}_{\tau_t}})^{-1}X^{\top}_{\mc{A}_{\tau_t}}(q + b\tau_{t+1} - q - b\tau_t) \quad  \text{using Equation (\ref{eq:parametrized_response_vector})} \\
    &= (X^{\top}_{\mc{A}_{\tau_t}}X_{\mc{A}_{\tau_t}})^{-1}X^{\top}_{\mc{A}_{\tau_t}}b(\tau_{t+1} - \tau_t)\\
    &= \nu_{\mathcal{A}_{\tau_t}}(\tau_{t+1} - \tau_t).
\end{align*}
where, we defined \(\nu_{\mathcal{A}_{\tau_t}}(\tau_t) =(X^{\top}_{\mc{A}_{\tau_t}}X_{\mc{A}_{\tau_t}})^{-1}X^{\top}_{\mc{A}_{\tau_t}}b \).
Similarly, for the non-active components, where \(\mcl{A}^c_{\tau_t} = \mcl{A}^c_{\tau_{t+1}}\) but, \(s_{\mathcal{A}^c_{\tau_t}}(\tau_t) \neq s_{\mathcal{A}^c_{\tau_{t}}}(\tau_{t+1})\),
\begin{align}
	-X^T_{\mathcal{A}^c_{\tau_t}}(y (\tau_t) - X_{\mathcal{A}_{\tau_t}}\beta_{\mathcal{A}_{\tau_t}}(\tau_{t})) 
    + \lambda s_{\mathcal{A}^c_{\tau_t}}(\tau_t) = 0 \label{eqn:lasso_normal_tau1_na}, \\ 
    -X^T_{\mathcal{A}^c_{\tau_{t}}}(y (\tau_{t+1}) - X_{\mathcal{A}_{\tau_{t}}}\beta_{\mathcal{A}_{\tau_{t}}}(\tau_{t+1})) 
    + \lambda s_{\mathcal{A}^c_{\tau_{t}}}(\tau_{t+1}) = 0 \label{eqn:lasso_normal_tau2_na}
\end{align}
%
%
Therefore, subtracting (\ref{eqn:lasso_normal_tau1_na}) from (\ref{eqn:lasso_normal_tau2_na}) we can write
\begin{align}
\lambda s_{\mathcal{A}^c_{\tau_{t}}}(\tau_{t+1}) - \lambda s_{\mathcal{A}^c_{\tau_t}}(\tau_t)(\tau_{t})
&= X^T_{\mathcal{A}^c_{\tau_t}}(b - X_{\mathcal{A}_{\tau_t}} \nu_{\mathcal{A}_{\tau_t}}(\tau_t))(\tau_{t+1} - \tau_{t}) \nonumber \\
   &= \gamma_{\mathcal{A}^c_{\tau_t}}(\tau_t) (\tau_{t+1} - \tau_{t}) \label{eq:gamma_appendix}
\end{align}
where we defined \(\gamma_{\mathcal{A}^c_{\tau_t}}(\tau_t) = X^T_{\mathcal{A}^c_{\tau_t}}(b - X_{\mathcal{A}_{\tau_t}} \nu_{\mathcal{A}_{\tau_t}}(\tau_t))\). 
\subsubsection{Derivation of step-size $\Delta_j$ in Equation (\ref{eq:step_length})}\label{LASSO:step-size_z_path}
\paragraph{Step-size of inclusion $\Delta_j^1$:}
Let's define \(\forall j \in \mcl{A}^c_{\tau_t}\),  \(c_j(\tau_t) =   x_j^{\top}\left( y(\tau_t) - X_{\mcl{A}_{\tau_t}}\beta_{\mcl{A}_{\tau_t}}(\tau_t) \right) \), then we can rewrite Equation (\ref{eq:opt_condn_shim_homotopy}) for non-active components as
\begin{equation}\label{eqn:tau_path_normal_eqn}
\begin{split}
    -c_j(\tau_t) + \lambda s_j(\tau_t) = 0\\
   \therefore \hspace{1cm} c_j(\tau_t) = \lambda s_j(\tau_t).
\end{split}
\end{equation}
Therefore, at any step \((\tau_{t+1} \rightarrow \tau_t + \Delta_j \)) any non-active feature \( (j \in \mathcal{A}^c_{\tau_t} ) \) becomes active when the following condition is satisfied. i.e.
\begin{equation}\label{eqn:tau_path_equi_corr_lamda}
    | c_j(\tau_{t+1}) | = \lambda.
\end{equation}
Now, let's consider a linear approximation of \(c_j(\tau_{t+1})\) by considering the value \(c_j(\tau_t)\) at $\tau_t$ i.e.
\begin{align}
    c_j(\tau_{t+1}) &= c_j(\tau_t) + (\tau_{t+1} - \tau_t) \frac{\partial c_j(\tau_t)}{\partial \tau} \nonumber \\
    &= c_j(\tau_t) + (\tau_{t+1} - \tau_t) g_j(\tau_t). \label{eq:lin_approx}
\end{align}
where, \(g_j(\tau_t) = \frac{\partial c_j(\tau_t)}{\partial \tau}\). 
By plugging (\ref{eq:lin_approx}) into  (\ref{eqn:tau_path_equi_corr_lamda})
and expanding (\ref{eqn:tau_path_equi_corr_lamda}) separately for positive and negative terms we can write the step-size of inclusion as 
\begin{equation*}
 \begin{split}
    \Delta_j^1  = \tau_{t+1} - \tau_t &= \underset{j \in \mathcal{A}_{\tau_t}^c}{\min} \Bigg( \frac{\lambda - c_j(\tau_t)}{g_j(\tau_t)}, \frac{-\lambda - c_j(\tau_t)}{g_j(\tau_t)} \Bigg)\\
    &= \underset{j \in \mathcal{A}_{\tau_t}^c}{\min} \Bigg( \frac{\pm \lambda - c_j(\tau_t)}{g_j(\tau_t)} \Bigg)\\
    &= \underset{j \in \mathcal{A}_{\tau_t}^c}{\min} \Bigg( \frac{\lambda \hspace{0.1cm} \text{sign}(g_j(\tau_t)) - c_j(\tau_t)}{g_j(\tau_t)} \Bigg)\\
    &= \underset{j \in \mathcal{A}_{\tau_t}^c}{\min} \Bigg( \lambda \frac{ \text{sign}(\gamma_j(\tau_t)) - s_j(\tau_t)}{\gamma_j(\tau_t)} \Bigg), \enspace \text{using } c_j(\tau_t) = \lambda s_j(\tau_t) \text{ in Equation }(\ref{eqn:tau_path_normal_eqn}).
\end{split}   
\end{equation*}
The last equality has been written considering the fact that \(g_j(\tau_t) = \gamma_j(\tau_t)\). The proof of this is given below.
\begin{proof}
\textnormal{We will now show that} \(g_j(\tau_t) = \gamma_j(\tau_t), \enspace \forall j \in \mathcal{A}^c_{\tau_t}\).
\textnormal{We know from (\ref{eq:gamma_appendix}) that \(\gamma_j(\tau_t), \enspace \forall j \in \mathcal{A}^c_{\tau_t}\), is}
\begin{equation*}
\begin{split}
    \gamma_j(\tau_t) &= \frac{\lambda s_j(\tau_{t+1}) - \lambda s_j(\tau_t)}{\tau_{t+1} - \tau_1},  \\
    &= \frac{c_j(\tau_{t+1}) - c_j(\tau_t)}{\tau_{t+1} - \tau_1}, \textnormal{ using } (\ref{eqn:tau_path_normal_eqn})\\
    &= \frac{\partial c_j(\tau_t)}{\partial \tau},\\
    &= g_j(\tau_t).
\end{split}
\end{equation*}
\end{proof}
%
%
%

\paragraph{Step-size of deletion $\Delta_j^2$:}
 A non zero variable becomes zero \ie
 \(\exists j \in \mcl{A}_{\tau_t} \text{ such that }: \beta_j(\tau_t) \neq 0 \text{ and } \beta_j(\tau_{t+1}) = 0 \enspace.\)
 \begin{align}\label{eq:lasso_step_size_tau_d2}
 \therefore \quad \beta_j(\tau_{t+1}) &= \beta_j(\tau_t) +  \Delta^2_j\nu_j(\tau_t) = 0 \Longrightarrow 
 \Delta^2_j = \min_{j \in \mcl{A}_{\tau_t}} \Bigg( - \frac{\beta_j(\tau_t)}{\nu_j(\tau_t)} \Bigg)_{++}.
 \end{align}

\subsubsection{Proofs of Lemmas \ref{lemma:lemma_1} and \ref{lemma:lemma_2} in  \S3.2.}

We first prove Lemma \ref{lemma:lemma_1}.
The pruning condition at any node $\ell$ in (\ref{eqn:pruning_condn}) is 
\begin{equation}\label{eqn:pruning_cond_node_l}
    |\rho_{\ell}(\tau_{t}, \tau_{t+1})| + \Delta_{\ell} |\eta_{\ell}(\tau_t)| < |\rho_k(\tau_{t}, \tau_{t+1})| - \Delta_{\ell} |\eta_k(\tau_t)|.
\end{equation}
Let $\Delta^\ast_\ell$ is the current minimum step-size, i.e. \(\Delta^\ast_\ell = \underset{t \in \{1, 2, \ldots, \ell\}}{\min}\{\Delta_t\} \). Now, if we consider the node $\ell$ to find the minimum step-size then, we are expecting that $\Delta_{\ell} \leq \Delta^\ast_{\ell}$.
Therefore, by construction, we can write
\begin{equation*}
\begin{split}
    |\rho_{\ell}(\tau_{t}, \tau_{t+1})| + \Delta_{\ell} |\eta_{\ell}(\tau_t)| &\leq |\rho_{\ell}(\tau_{t}, \tau_{t+1})| + \Delta^\ast_{\ell} |\eta_{\ell}(\tau_t)| \quad \text{and,}\\
    |\rho_k(\tau_{t}, \tau_{t+1})| - \Delta^\ast_{\ell} |\eta_k(\tau_t)| &\leq |\rho_k(\tau_{t}, \tau_{t+1})| - \Delta_{\ell} |\eta_k(\tau_t)|.
\end{split}
\end{equation*}
Therefore, (\ref{eqn:pruning_cond_node_l}) is equivalent to %
\begin{equation}\label{eqn:pruning_cond_node_l_opt_step}
    |\rho_{\ell}(\tau_t, \tau_{t + 1})| + \Delta^\ast_\ell |\eta_\ell(\tau_t)| <  |\rho_k(\tau_t, \tau_{t + 1})| - \Delta^\ast_\ell |\eta_k(\tau_t)|.
\end{equation}
This completes the proof of Lemma \ref{lemma:lemma_1}. Therefore, Lemma \ref{lemma:lemma_1} is the new pruning condition using the current minimum step-size, i.e  $\Delta^\ast_\ell$. 
Note that we can further simplify Lemma \ref{lemma:lemma_1} as follows. We can write 
\begin{equation}\label{eqn:rho_tau_t_plus_1}
\begin{split}
\rho_{\ell}(\tau_t, \tau_{t + 1}) &= |x_{\ell}^{\top}\big(y(\tau_{t+1}) -  X_{\mc{A}_{\tau_t}}\beta_{\mc{A}_{\tau_t}}(\tau_t)\big)| \\
   &= |x_{\ell}^{\top}\big(y(\tau_{t}) + \Delta^\ast_{\ell}b) -  X_{\mc{A}_{\tau_t}}\beta_{\mc{A}_{\tau_t}}(\tau_t)\big)| \textnormal{ using (\ref{eq:parametrized_response_vector})}\\
   &= |x_{\ell}^{\top}\big(y(\tau_{t}) -  X_{\mc{A}_{\tau_t}}\beta_{\mc{A}_{\tau_t}}(\tau_t)\big) + \Delta^\ast_{\ell}x_{\ell}^{\top}b|\\
   &= |\rho_{\ell}(\tau_t) + \Delta^\ast_{\ell}\theta_\ell|,
\end{split}
\end{equation}
where $\theta_\ell = x_{\ell}^{\top}b$ and $\rho_\ell(\tau_t) = x_{\ell}^{\top}\big(y(\tau_{t}) -  X_{\mc{A}_{\tau_t}}\beta_{\mc{A}_{\tau_t}}(\tau_t)\big) $.
We know that
\begin{equation}\label{eqn:traingular_inequality}
\begin{split}
    |\rho_{\ell}(\tau_t) + \Delta^\ast_{\ell}\theta_\ell| &\geq |\rho_{\ell}(\tau_t)| - \Delta^\ast_{\ell}|\theta_\ell| \quad \text{and,}\\
    |\rho_{\ell}(\tau_t) + \Delta^\ast_{\ell}\theta_\ell| &\leq |\rho_{\ell}(\tau_t)| + \Delta^\ast_{\ell}|\theta_\ell|.
\end{split}
\end{equation}
Now using (\ref{eqn:rho_tau_t_plus_1}) and (\ref{eqn:traingular_inequality}) we can further write (\ref{eqn:pruning_cond_node_l_opt_step}) as
\begin{align}\label{eqn:pruning_cond_node_l_opt_step2}
    |\rho_{\ell}(\tau_t)| + \Delta^{\ast}_{\ell} |\theta_\ell| + \Delta^\ast_\ell |\eta_\ell(\tau_t)| &<  |\rho_k(\tau_t)| - \Delta^\ast_{\ell} |\theta_k| - \Delta^\ast_{\ell} |\eta_k(\tau_t)|.
\end{align}
Therefore, (\ref{eqn:pruning_cond_node_l_opt_step2}) serves as the simplified expression of the Lemma \ref{lemma:lemma_1}. 
Next, we provide two propositions which we use to prove Lemma \ref{lemma:lemma_2}.

\begin{proposition} \label{prop_1} (Tree anti-monotonicity)
A tree is constructed in such a way that for any pair of nodes (\(\ell, \ell^\prime)\), where $\ell$ is the ancestor of $\ell^\prime$, i.e., \(\ell^\prime \supset \ell \), the following conditions are satisfied 
 \begin{equation}\label{eq:tree_anti_monotonicity_prop}
    \hst  x_{i \ell^\prime} = 1 \implies x_{i\ell } = 1  \hst
 \text{ and conversely,} \hst
     x_{i\ell } = 0 \implies x_{i \ell^\prime} = 0  \quad \forall i \in [n].
 \end{equation}
\end{proposition}

\begin{proposition}\label{prop_2}
If Proposition 1 holds, then $\forall \ell^\prime \supset \ell $, we have 
\begin{align*}
	\lvert \rho_\ell (\tau_t) \lvert &\geq \lvert \rho_{\ell^\prime} (\tau_t) \lvert, \\ 
	\lvert \eta_\ell (\tau_t) \lvert &\geq \lvert \eta_{\ell^\prime} (\tau_t) \lvert, \\ 
	|\theta_\ell  |  &\geq |\theta_{\ell^\prime}|.
\end{align*}
\end{proposition}

\paragraph{Proof for Proposition 2:}
If Proposition 1 holds, we have 
\begin{align*}
	\lvert \rho_\ell (\tau_t) \lvert &= \lvert x_\ell^\top (y(\tau_t) - X_{\cA_{\tau_t}} \beta_{\cA_{\tau_t}}(\tau_t)), \\ 
	& = \lvert x_\ell^\top w(\tau_t) \lvert, \\ 
	& \geq \lvert x_{\ell^\prime}^\top w(\tau_t) \lvert \quad \text{ if }~  w(\tau_t) \geq 0,\\
	& =: \lvert \rho_{\ell^\prime} (\tau_t) \lvert,
\end{align*}
where $w(\tau_t) = y(\tau_t) - X_{\cA_{\tau_t}} \beta_{\cA_{\tau_t}}(\tau_t)$.
Similarly, we also have
\begin{align*}
	\lvert \eta_\ell (\tau_t) \lvert &= \lvert x_\ell^\top X_{\cA_{\tau_t}} \nu_{\cA_{\tau_t}} (\tau_t) \lvert, \\ 
	& = \lvert x_\ell^\top v(\tau_t) \lvert, \\ 
	& \geq \lvert x_{\ell^\prime}^\top v(\tau_t) \lvert \quad \text{ if }~  v(\tau_t) \geq 0,\\
	& =: \lvert \eta_{\ell^\prime} (\tau_t) \lvert,
\end{align*}
where $v(\tau_t) = X_{\cA_{\tau_t}} \nu_{\cA_{\tau_t}} (\tau_t)$, and
\begin{align*}
	\lvert \theta_\ell  \lvert &= \lvert x_\ell^\top b \lvert, \\ 
	& \geq \lvert x_{\ell^\prime}^\top b \lvert \quad \text{ if }~  b \geq 0,\\
	&=: \lvert \theta_{\ell^\prime}  \lvert,
\end{align*}
This completes the proof of Proposition 2.

Proposition 2 will be used to prove Lemma \ref{lemma:lemma_2}.
We will prove Lemma \ref{lemma:lemma_2} by contradiction i.e. we assume that (\ref{eqn:pruning_cond_node_l_opt_step2}) holds and \(\forall \ell^{\prime} \text{ s.t. } \ell^{\prime} \supset \ell\),  $\Delta_{\ell^{\prime}} < \Delta_l^{\ast}$.
\begin{equation*}
    \begin{split}
    \therefore \quad |\rho_k(\tau_t)| - \Delta_{\ell^{\prime}} |\theta_k| - \Delta_{\ell^{\prime}} |\eta_k(\tau_t)| &> |\rho_k(\tau_t)| - \Delta^\ast_{\ell} |\theta_k| - \Delta^\ast_{\ell} |\eta_k(\tau_t)|, \hst \because \Delta_{\ell^{\prime}} < \Delta_l^{\ast}\\
    &> |\rho_{\ell}(\tau_t)| + \Delta^\ast_{\ell} |\theta_\ell| + \Delta^\ast_{\ell} |\eta_{\ell}(\tau_t)|, \hst \text{using } (\ref{eqn:pruning_cond_node_l_opt_step2})\\
    &> |\rho_{\ell^{\prime}}(\tau_t)| + \Delta^\ast_{\ell} |\theta_{\ell^\prime}| + \Delta^\ast_{\ell} |\eta_{\ell^{\prime}}(\tau_t)|, \hst \text{using Proposition 2,}\\
    &> |\rho_{\ell^{\prime}}(\tau_t)| + \Delta_{\ell^{\prime}} |\theta_{\ell^\prime}| + \Delta_{\ell^{\prime}} |\eta_{\ell^{\prime}}(\tau_t)|, \hst \because \Delta_{\ell^{\prime}} < \Delta_l^{\ast}.
    \end{split}
\end{equation*}
Therefore, we got
\begin{equation*}
\begin{split}
  &\therefore \quad |\rho_k(\tau_t)| - \Delta_{\ell^{\prime}} |\theta_k| - \Delta_{\ell^{\prime}} |\eta_k(\tau_t)| > |\rho_{\ell^{\prime}}(\tau_t)| + \Delta_{\ell^{\prime}} |\theta_{\ell^\prime}| + \Delta_{\ell^{\prime}} |\eta_{\ell^{\prime}}(\tau_t)|\\
  &\implies \ell^{\prime} \text{ is infeasible} \hst (\text{using } (\ref{eqn:pruning_cond_node_l_opt_step2})) \implies \Delta_{\ell^{\prime}} \nless \Delta_{\ell}^{\ast}.
\end{split}
\end{equation*}
This completes the proof of Lemma \ref{lemma:lemma_2}.

If any of $w(\tau_t)$, $v(\tau_t)$ and $b$ in Proposition 2 contains at least one negative element, then we can no longer use Lemma 2.
Hence, using the following Proposition 3, we can propose Lemma 3 as a general pruning condition.

\begin{proposition} We can write
\begin{align*}
	|\rho_\ell (\tau_t)| & \leq b_{\ell, w(\tau_t)}, \\ 
	|\eta_\ell (\tau_t)| & \leq b_{\ell, v(\tau_t)}, \\ 
	|\theta_\ell| & \leq b_{\ell, \theta},
\end{align*}
where 
\begin{align*}
 	 b_{\ell, w(\tau_t)} &= \max \big\{ \sum_{w_i(\tau_t) < 0} \lvert w_i(\tau_t) \lvert x_{i\ell}, \sum_{w_i(\tau_t) > 0} \lvert w_i(\tau_t) \lvert x_{i\ell} \big\}  \\ 
	 b_{\ell, v(\tau_t)} &= \max \big\{ \sum_{v_i(\tau_t) < 0} \lvert v_i(\tau_t) \lvert x_{i\ell}, \sum_{v_i(\tau_t) > 0} \lvert v_i(\tau_t) \lvert x_{i\ell} \big\} \\ 
	 b_{\ell, \theta} &= \max \big\{ \sum_{b_i < 0} \lvert b_i \lvert x_{i\ell}, \sum_{b_i > 0} \lvert b_i \lvert x_{i\ell} \big\}.
\end{align*}

\end{proposition}

\paragraph{Proof of Proposition 3:}
We have 
\begin{align*}
	|\rho_\ell (\tau_t)| &= |x_\ell^\top w(\tau_t)| \\ 
	&=  \left | \sum 
	\limits_{i=1}^n w_{i\ell} x_{i \ell} \right |  \\ 
	&=  \left | \sum 
	\limits_{w_{i\ell} > 0} |w_{i\ell}| x_{i \ell} - \sum  \limits_{w_{i\ell} < 0} |w_{i\ell}| x_{i \ell}\right | \\ 
	&\leq  \max \left \{ \sum 
	\limits_{w_{i\ell} > 0} |w_{i\ell}| x_{i \ell}, \sum  \limits_{w_{i\ell} < 0} |w_{i\ell}| x_{i \ell} \right \} =: b_{\ell, w(\tau_t)}.
\end{align*}
Similarly, 
\begin{align*}
	|\eta_\ell (\tau_t)| &= |x_\ell^\top v(\tau_t)| \\ 
	&=  \left | \sum 
	\limits_{i=1}^n v_{i\ell} x_{i \ell} \right |  \\ 
	&=  \left | \sum 
	\limits_{v_{i\ell} > 0} |v_{i\ell}| x_{i \ell} - \sum  \limits_{v_{i\ell} < 0} |v_{i\ell}| x_{i \ell}\right | \\ 
	&\leq  \max \left \{ \sum 
	\limits_{v_{i\ell} > 0} |v_{i\ell}| x_{i \ell}, \sum  \limits_{v_{i\ell} < 0} |v_{i\ell}| x_{i \ell} \right \} =: b_{\ell, v(\tau_t)}
\end{align*}
and
\begin{align*}
	|\theta_\ell| &= |x_\ell^\top b| \\ 
	&=  \left | \sum 
	\limits_{i=1}^n b_i x_{i \ell} \right |  \\ 
	&=  \left | \sum 
	\limits_{b_i > 0} |b_i| x_{i \ell} - \sum  \limits_{b_i < 0} |b_i| x_{i \ell}\right | \\ 
	&\leq  \max \left \{ \sum 
	\limits_{b_i > 0} |b_i| x_{i \ell}, \sum  \limits_{b_i < 0} |b_i| x_{i \ell} \right \} =: b_{\ell, \theta}.
\end{align*}
This completes the proof of Proposition 3.

\begin{lemma} Using Proposition 3 we can show that
$\forall \ell^\prime \supset \ell$, if 
\begin{align} \label{eq:lemma_3_eq}
	b_{\ell, w(\tau_t)} + \Delta_\ell^\ast b_{\ell, \theta} + \Delta_\ell^\ast b_{\ell, v(\tau_t)} < |\rho_k(\tau_t)| - \Delta^\ast_\ell |\theta_k| - \Delta_\ell^\ast |\eta_k(\tau_t)|,
\end{align}
then $\Delta_{\ell^\prime} > \Delta_\ell^\ast$.
\end{lemma}

Before proving Lemma 3, we introduce Proposition 4 which will be used to prove Lemma 3:

\begin{proposition}
If Proposition 1 holds, we have 
\begin{align*}
	b_{\ell, w(\tau_t)} & \geq b_{\ell^\prime, w(\tau_t)}, \\ 
	b_{\ell, v(\tau_t)} & \geq b_{\ell^\prime, v(\tau_t)}, \\ 
	b_{\ell, \theta} & \geq b_{\ell^\prime, \theta}.
\end{align*}
\end{proposition}

\paragraph{Proof of Proposition 4:}
If Proposition 1 holds, we have 
\begin{align*}
	b_{\ell, w(\tau_t)} &= \max \big\{ \sum_{w_i(\tau_t) < 0} \lvert w_i(\tau_t) \lvert x_{i\ell}, \sum_{w_i(\tau_t) > 0} \lvert w_i(\tau_t) \lvert x_{i\ell} \big\} \\ 
	&\geq \max \big\{ \sum_{w_i(\tau_t) < 0} \lvert w_i(\tau_t) \lvert x_{i\ell^\prime}, \sum_{w_i(\tau_t) > 0} \lvert w_i(\tau_t) \lvert x_{i\ell^\prime} \big\} =: b_{\ell^\prime, w(\tau_t)}.
\end{align*} 
Similarly, we also have
\begin{align*}
	b_{\ell, v(\tau_t)} &= \max \big\{ \sum_{v_i(\tau_t) < 0} \lvert v_i(\tau_t) \lvert x_{i\ell}, \sum_{v_i(\tau_t) > 0} \lvert v_i(\tau_t) \lvert x_{i\ell} \big\} \\ 
	&\geq \max \big\{ \sum_{v_i(\tau_t) < 0} \lvert v_i(\tau_t) \lvert x_{i\ell^\prime}, \sum_{v_i(\tau_t) > 0} \lvert v_i(\tau_t) \lvert x_{i\ell^\prime} \big\} =: b_{\ell^\prime, v(\tau_t)}.
\end{align*}
and
\begin{align*}
	b_{\ell, \theta} &= \max \big\{ \sum_{b_i < 0} \lvert b_i \lvert x_{i\ell}, \sum_{b_i > 0} \lvert b_i \lvert x_{i\ell} \big\} \\ 
	&\geq \max \big\{ \sum_{b_i < 0} \lvert b_i \lvert x_{i\ell^\prime}, \sum_{b_i > 0} \lvert b_i \lvert x_{i\ell^\prime} \big\} =: b_{\ell^\prime, \theta}.
\end{align*}

\paragraph{Proof of Lemma 3:}

%
%

%
We will prove Lemma 3 by contradiction i.e. we assume that (\ref{eq:lemma_3_eq}) holds and \(\forall \ell^{\prime} \text{ s.t. } \ell^{\prime} \supset \ell\),  $\Delta_{\ell^{\prime}} < \Delta_l^{\ast}$.
\begin{equation*}
    \begin{split}
    \therefore \quad |\rho_k(\tau_t)| - \Delta_{\ell^{\prime}} |\theta_k| - \Delta_{\ell^{\prime}} |\eta_k(\tau_t)| &> |\rho_k(\tau_t)| - \Delta^\ast_{\ell} |\theta_k| - \Delta^\ast_{\ell} |\eta_k(\tau_t)|, \hst \because \Delta_{\ell^{\prime}} < \Delta_l^{\ast}\\
    &> b_{\ell, w(\tau_t)} + \Delta_\ell^\ast b_{\ell, \theta} + \Delta_\ell^\ast b_{\ell, v(\tau_t)}, \hst \text{using } (\ref{eq:lemma_3_eq})\\
    &> b_{\ell^\prime, w(\tau_t)} + \Delta_{\ell}^\ast b_{\ell^\prime, \theta} + \Delta_\ell^\ast b_{\ell^\prime, v(\tau_t)},  \hst \text{using Proposition 3,}\\
    &> b_{\ell^\prime, w(\tau_t)} + \Delta_{\ell^\prime} b_{\ell^\prime, \theta} + \Delta_{\ell^\prime}b_{\ell^\prime, v(\tau_t)}, \hst \because \Delta_{\ell^{\prime}} < \Delta_l^{\ast}.
    \end{split}
\end{equation*}
Therefore, we got
\begin{equation*}
\begin{split}
  &|\rho_k(\tau_t)| - \Delta_{\ell^{\prime}} |\theta_k| - \Delta_{\ell^{\prime}} |\eta_k(\tau_t)| > b_{\ell^\prime, w(\tau_t)} + \Delta_{\ell^\prime} b_{\ell^\prime, \theta} + \Delta_{\ell^\prime}b_{\ell^\prime, v(\tau_t)}\\
  &\implies \ell^{\prime} \text{ is infeasible} \hst (\text{using } (\ref{eq:lemma_3_eq})) \implies \Delta_{\ell^{\prime}} \nless \Delta_{\ell}^{\ast}.
\end{split}
\end{equation*}
This completes the proof of Lemma 3.

%
%
Hence, if the pruning condition in Lemma 3 holds, then we do not need to search the sub-tree with $\ell$ as the root node, and hence increasing the efficiency of the search procedure [\cite{ERP_Tsuda}].
%
%
%

\subsection{LASSO: $\lambda$-path}
\subsubsection{$\lambda$-path: path \wrt to $\lambda$ ($\tau$ fixed)}
Since $\tau$ is fixed, we drop it from the notation. The normal equation of the $\lambda$-path can be written as 
\begin{align*}
    - X^T\left( y - X\beta \right) + \lambda s(\lambda) &= 0
\end{align*}
where, \(s(\lambda)\) is the sub-differential defined as %
\begin{equation*}\label{eq:sub-diff_lambda_LASSO}
    s_j(\lambda) \in \begin{cases}
    \{-1, +1\}, \enspace \text{if} \enspace \beta_j(\lambda) \neq 0\\
    [-1, +1],  \enspace \enspace \text{if} \enspace \beta_j(\lambda) = 0.
    \end{cases}
\end{equation*}
Now, if we consider two $\lambda$ values (\(\lambda_t > \lambda_{t+1}\)) at which the active set does not change (i.e. $\mcl{A}_{\lambda_t} = \mcl{A}_{\lambda_{t+1}}$) and the sign of the active coefficients also remain the same (i.e. $s_{\mc{A}_{\lambda_t}}(\lambda_t) = s_{\mc{A}_{\lambda_{t+1}}}(\lambda_{t+1})$) , then we can write
\begin{equation}\label{eqn:lasso_lmd_path_beta_piecewise_linear}
    \beta_{\mc{A}_{\lambda_{t}}}(\lambda_{t+1}) - \beta_{\mc{A}_{\lambda_t}}(\lambda_t) = - \nu_{\mc{A}_{\lambda_t}}(\lambda_t) (\lambda_{t+1} - \lambda_t)
\end{equation}
where, \(\nu_{\mc{A}_{\lambda_t}}(\lambda_t) = (X^T_{\mc{A}_{\lambda_t}}X_{\mc{A}_{\lambda_t}})^{-1}s_{\mc{A}_{\lambda_t}}(\lambda_t)\). The derivation of \(\nu_{\mc{A}_{\lambda_t}}(\lambda_t)\) is given in \ref
{LASSO:direction_vector_lambda_path}. Note that \(\nu_{\mathcal{A}_{\lambda}}(\lambda)\)  is constant  for all real values of \(\lambda \in [\lambda_t, \lambda_{t+1})\) and thus, equation (\ref{eqn:lasso_lmd_path_beta_piecewise_linear}) states that \(\beta(\lambda)\) is piece-wise linear in \(\lambda\) for a fixed \(\tau\). To draw the curve of solutions as a function of $\lambda$, we need to check when the active set changes. If $\lambda_{t+1}$ is the next zero crossing point then either of the following two events happens.

$\bullet$ A zero variable becomes non-zero \ie
\( \exists j \in A^c_{\lambda_t} \text{ s.t. }: |x_{j}^{\top}(y - X_{\mcl{A}_{\lambda_t}}\beta_{\mcl{A}_{\lambda_t}}(\lambda_{t+1}))| = \lambda_{t+1} \text{ or,}\)

$\bullet$ A non zero variable becomes zero \ie \(\exists j \in \mcl{A}_{\lambda_t} \text{ s.t.}:\beta_j(\lambda_t) \neq 0 \text{ but } \beta_j(\lambda_{t+1}) = 0 \enspace.\)

Overall, the next change of the active set happens at \(\lambda_{t+1} = \lambda_t - \Delta_j, \) where
\begin{equation}\label{eq:lasso_lmd_path_step_length}
    \Delta_j = min(\Delta_j^1, \Delta_j^2) = \min\left(\min_{j \in \mc{A}^c_{\lambda_t}} \Big( \frac{(x_k \pm x_j)^Tw(\lambda_t)}{(x_k \pm x_j)^Tv(\lambda_t)} \Big)_{++}, \hst \min_{j \in \mcl{A}_{\tau_t}} \Big( - \frac{\beta_j(\lambda_t)}{\nu_j(\lambda_t)} \Big)_{++} \right) \enspace.
\end{equation}
where, \(w(\lambda_t) = y - X_{\mcl{A}_{\lambda_t}}\beta_{\mcl{A}_{\lambda_t}}(\lambda_t)\), and \(v(\lambda_t) = X_{\mcl{A}_{\lambda_t}}\nu_{\mcl{A}_{\lambda_t}}(\lambda_t)\). The derivation of the step-size of inclusion ($\Delta_j^1$) is given in \ref{LASSO:step-size_lmd_path_d1}.

%
%
%
\subsubsection{Direction vector ($\lambda$-path)}\label{LASSO:direction_vector_lambda_path}
Let's consider the normal equation at $\lambda_t$ and $\lambda_{t+1}$ for the active components
\begin{equation}\label{eqn:normal_lambda1}
    -X^{\top}_{\mathcal{A}_{\lambda_t}}(y - X_{\mathcal{A}_{\lambda_t}}\beta_{\mathcal{A}_{\lambda_t}}(\lambda_{t})) 
    + \lambda_t s_{\mathcal{A}_{\lambda_t}}(\lambda_{t}) = 0,
\end{equation}
\begin{equation}\label{eqn:normal_lambda2}
    -X^{\top}_{\mathcal{A}_{\lambda_{t}}}(y - X_{\mathcal{A}_{\lambda_{t}}}\beta_{\mathcal{A}_{\lambda_{t}}}(\lambda_{t+1})) 
    + \lambda_{t} s_{\mathcal{A}_{\lambda_{t}}}(\lambda_{t}) = 0.
\end{equation}
Note that \(\mcl{A}_{\lambda_t} = \mcl{A}_{\lambda_{t+1}}\) and \(s_{\mc{A}_{\lambda_t}}(\lambda_{t}) = s_{\mcl{A}_{\lambda_{t}}}(\lambda_{t+1})\). Therefore, subtracting (\ref{eqn:normal_lambda1}) from (\ref{eqn:normal_lambda2}) one can write
\begin{align*}
    \beta_{\mathcal{A}_{\lambda_{t}}}(\lambda_{t+1}) - \beta_{\mathcal{A}_{\lambda_t}}(\lambda_{t}) = -\nu_{\mathcal{A}_{\lambda_t}}(\lambda_{t}) (\lambda_{t+1} - \lambda_t),
\end{align*}
where, we defined \(\nu_{\mathcal{A}_{\lambda_t}}(\lambda_{t}) = (X^{\top}_{\mathcal{A}_{\lambda_t}}X_{\mathcal{A}_{\lambda_t}})^{-1}s_{\mathcal{A}_{\lambda_t}}(\lambda_t)\).
%
\subsubsection{Step-size of inclusion ($\lambda$-path)}\label{LASSO:step-size_lmd_path_d1}
The optimality condition for the active features \(j \in \mathcal{A}_{\lambda_t}\) of the $\lambda$-path of LASSO can be written as
\begin{align*}
    - x_j^{\top}\left( y - X_{\mcl{A}_{\lambda_t}}\beta_{\mcl{A}_{\lambda_t}}(\lambda_t) \right) + \lambda s(\beta_j) &= 0,\\
    x_j^{\top}\left( y - X_{\mcl{A}_{\lambda_t}}\beta_{\mcl{A}_{\lambda_t}}(\lambda_t) \right) &= \lambda s_j,\\
    \therefore \hspace{1cm} \left| x_j^{\top}\left( y - X_{\mcl{A}_{\lambda_t}}\beta_{\mcl{A}_{\lambda_t}}(\lambda_t) \right) \right| &= \lambda.
\end{align*}
Therefore, at any step \((\lambda_t \rightarrow \lambda_t - \Delta_j) \) any non-active feature \( (j \in \mathcal{A}^c_{\lambda_t} ) \) becomes active when the following condition is satisfied i.e.
\begin{align*}
    \left| x_j^{\top}\left( y - X_{\mcl{A}_{\lambda_t}}(\beta_{\mcl{A}_{\lambda_t}}(\lambda_t) + \Delta_j\nu(\lambda_t)) \right) \right| &= \left| x_k^{\top}\left(y - X_{\mcl{A}_{\lambda_t}}(\beta_{\mcl{A}_{\lambda_t}}(\lambda_t) + \Delta_j\nu(\lambda_t)) \right) \right|, \hst \forall j \in \mathcal{A}^c_{\lambda_t}, \forall k \in \mathcal{A}_{\lambda_t}\\
    \left|x_j^{\top}(w(\lambda_t) - \Delta_j X_{\mcl{A}_{\lambda_t}}v(\lambda_t)) \right| &= \left|x_k^{\top}(w(\lambda_t) - \Delta_j X_{\mcl{A}_{\lambda_t}}v(\lambda_t)) \right|,
\end{align*}
where \(w(\lambda_t) = y - X_{\mcl{A}_{\lambda_t}}\beta_{\mcl{A}_{\lambda_t}}(\lambda_t) \) and  \(v(\lambda_t) = X_{\mcl{A}_{\lambda_t}}\nu_{\mcl{A}_{\lambda_t}}(\lambda_t) \). Now, considering the positive and negative terms separately one can write the step-size of inclusion as -
\begin{align*}
    \Delta_j^1 &= \underset{j \in \mathcal{A}^c_{\lambda_t}}{min} \Bigg( \frac{(x_k \pm x_j)^{\top}w(\lambda_t)}{(x_k \pm x_j)^{\top}v(\lambda_t)} \Bigg).
\end{align*}
\subsubsection{Tree pruning ($\lambda$-path)}
The derivation of this pruning condition is also given in \citep{ERP_Tsuda}. However, here we provide the same derivation in our notation to make it self-contained. 
Similar to (\ref{eqn:pruning_condn}), the pruning condition of the $\lambda$-path can be written as 
\begin{equation}\label{eq:lasso_tree_pruning_lmd_path1}
    \lvert \rho_{\ell} (\lambda_t) \lvert + \Delta^{\ast}_{\ell} \lvert \eta_{\ell}(\lambda_t) \lvert < \lvert \rho_k(\lambda_t) \lvert - \Delta^{\ast}_{\ell} \lvert \eta_k(\lambda_t) \lvert,
\end{equation}
where \(\rho_{\ell}(\lambda_t) = x_{\ell}^{\top}w(\lambda_t) \) and \(\eta_{\ell}(\lambda_t) =x_{\ell}^{\top} v(\lambda_t) \quad \forall \ell \in \mathcal{A}^c_{\lambda_t} \), \(\rho_k(\lambda_t) = x_{k}^{\top}w(\lambda_t) \) and \(\eta_k(\lambda_t) = x_k^{\top} v(\lambda_t),  \quad \forall k \in \mathcal{A}_{\lambda_t} \) . 
Now similar to the Proposition 3 we can also write
\begin{proposition} \label{proposition:lasso_tree_antimonotonicity_lmd_path}
\textnormal{Using the tree anti-monotonicity property (\ref{eq:tree_anti_monotonicity_prop}) we can easily show that $\forall \ell^{\prime} $ s.t. $\ell^{\prime} \supset \ell$, the following conditions are satisfied i.e.} 
\begin{align*}
 | \rho_{\ell^{\prime}} (\lambda_t) | = \lvert \sum_i w_i(\lambda_t) x_{i\ell^{\prime}} \lvert &\leq max \big\{ \sum_{w_i(\lambda_t) < 0} \lvert w_i(\lambda_t) \lvert x_{i\ell}, \sum_{w_i(\lambda_t) > 0} \lvert w_i(\lambda_t) \lvert x_{i\ell} \big\} :=  b_{w(\lambda_t)},\\
|\eta_{\ell^{\prime}}(\lambda_t)| = \lvert \sum_i v_i(\lambda_t) x_{i\ell^{\prime}} \lvert &\leq max \big\{ \sum_{v_i(\lambda_t) < 0} \lvert v_i(\lambda_t) \lvert x_{i\ell}, \sum_{v_i(\lambda_t) > 0} \lvert v_i(\lambda_t) \lvert x_{i\ell} \big\} := b_{v(\lambda_t)}.
\end{align*}
\end{proposition}
Therefore, similar to the $\tau$-path (\ref{eq:lemma_3_eq}) the pruning condition of the $\lambda$-path (\ref{eq:lasso_tree_pruning_lmd_path1}) can be written as
\begin{equation}\label{eq:lasso_pruning_cond_lmd_path3}
    b_{w(\lambda_t)} + \Delta^{\ast}_{\ell}  b_{v(\lambda_t)} <  \lvert \rho_k(\lambda_t)| - \Delta^{\ast}_{\ell} \lvert \eta_k(\lambda_t) \lvert.
\end{equation}
%
%
%
%
%

The complete algorithm for the selection path ($\lambda$-path) is given in Algorithm\ref{algo:selection_path}. 
\begin{algorithm}[H]
\caption{$\lambda$-path}
\label{algo:selection_path}
\begin{algorithmic}[1]
\State \textbf{Input:} \(Z, y, \lambda \).
\State Initialization: \( \lambda_t = \lambda_{max} = ||X^{\top}y||_{\infty}, \hst \mcl{A}_{\lambda_t} = \underset{j}{\textit{arg max}} |X^{\top}y|_j, \quad \beta(\lambda_t)=0,\)
\State \(\nu_{\mc{A}_{\lambda_t}}(\lambda_t) = (X_{\mc{A}_{\lambda_t}}^{\top}X_{\mc{A}_{\lambda_t}})^{-1} sign(\beta_{\mc{A}_{\lambda_t}}(\lambda_t))\), \hst \(\nu_{\mc{A}^c_{\lambda_t}}(\lambda_t) = 0\).
\State \textbf{while}  \(\lambda_t \geq \lambda \hs \textbf{do}\). 
 %
    
    
    \State Compute step-length \(\Delta_j \leftarrow \) Equation (\ref{eq:lasso_lmd_path_step_length}).
    
    \State If \(\Delta_j = \Delta_j^1\), add $j$ into \(\mcl{A}_{\lambda_t}\). \Comment{Inclusion}
    
    \State If, \(\Delta_j = \Delta_j^2\), remove $j$ from \(\mcl{A}_{\lambda_t}\). \Comment{Deletion}.
    
    \State Update: \( \lambda_t \leftarrow \lambda_t - \Delta_j, \beta_{\mcl{A}_{\lambda_t}}(\lambda_{t+1}) \leftarrow \beta_{\mcl{A}_{\lambda_t}}(\lambda_t) + \Delta_j \nu_{\mcl{A}_{\lambda_t}}(\lambda_t), \newline
    \nu_{\mc{A}_{\lambda_{t+1}}}(\lambda_{t+1}) = (X_{\mc{A}_{\lambda_{t+1}}}^{\top}X_{\mc{A}_{\lambda_{t+1}}})^{-1} sign(\beta_{\mc{A}_{\lambda_{t+1}}}(\lambda_{t+1})), \quad \nu_{\mc{A}^c_{\lambda_{t+1}}}(\lambda_{t+1}) = 0. \)
    
    
    
    \State \textbf{end while}
    \State \textbf{Output:} \( \beta(\lambda), \mathcal{A}_{\lambda} \).
\end{algorithmic}
\end{algorithm}

\section{Extension for Elastic Net (ElNet)}
A common problem of the LASSO is that if the data has correlated features then, the LASSO picks only one of them and ignores the rest, which leads to instability. To solve this problem \cite{zou2005regularization} proposed the Elastic Net (ElNet). This feature correlation problem is very much evident in SHIM type problem, and hence we extended our framework for the Elastic Net. To extend our framework for the Elastic Net, we need to solve the following optimization problem.
\begin{equation}\label{obj:primal_elnet}
    \beta(\lambda, \tau) \in \argmin_{\beta \in \bbR^p} \frac{1}{2}\norm{y(\tau) - X\beta}_2^2 + \frac{1}{2}\alpha \norm{\beta}_2^2  + \lambda \norm{\beta}_1.
\end{equation}
\subsection{$\lambda$-path: path \wrt to $\lambda$ ($\tau$ fixed)}
Similar to the LASSO, the normal equation can be written as
\begin{align*}\label{eqn:ElNet_lmd_path_normal_eqn}
    - X^{\top}\left( y - X\beta(\lambda) \right) + \alpha \beta(\lambda) + \lambda s(\lambda) &= 0.
\end{align*}
where, \(s(\lambda)\) is the sub-differential that can be defined in a similar fashion as done in the case of the $\lambda$-path for the LASSO (\ref{eq:sub-diff_lambda_LASSO}). Now, if we consider two $\lambda$ values (\(\lambda_t > \lambda_{t+1}\)) at which the active set does not change (i.e. $\mathcal{A}_{\lambda_t} = \mathcal{A}_{\lambda_{t+1}}$) and the sign of the active coefficients also remain the same (i.e. $s_{\mathcal{A}_{\lambda_t}}(\lambda_t) = s_{\mathcal{A}_{\lambda_{t}}}(\lambda_{t+1})$) , then we can write
\begin{align}
    \beta_{\mathcal{A}_{\lambda_{t}}} (\lambda_{t + 1}) - \beta_{\mathcal{A}_{\lambda_t}} (\lambda_t) &= -\nu_{\mathcal{A}_{\lambda_t}} (\lambda_t) (\lambda_{t+1} - \lambda_t),
\end{align}
where, \(\nu_{\mathcal{A}_{\lambda_t}}(\lambda_t) = (X^{\top}_{\mathcal{A}_{\lambda_t}}X_{\mathcal{A}_{\lambda_t}} + \alpha I_{|\mathcal{A}_{\lambda_t}|})^{-1}s_{\mathcal{A}_{\lambda_t}}(\lambda_t)\). Note that here the only change in the direction vectors compared to the LASSO is the addition of an \(\alpha I_{|\mathcal{A}_{\lambda_t}|}\) term to the expression of \(\nu_{\mathcal{A}_{\lambda_t}}(\lambda_t)\). Now, similar to the LASSO we can derive the step-size of deletion ($\Delta_j^2$) considering this updated expression of the direction vector. However, to derive the step-size of inclusion ($\Delta_j^1$), we need a different approach. The elastic net optimization problem can actually be formulated as a LASSO optimization problem using augmented data.
If we consider an augmented data defined as \(\Tilde{X} = \begin{pmatrix} X\\ \sqrt{\alpha} I_p \end{pmatrix} \) and \(\Tilde{y} = \begin{pmatrix}y \\ 0 \end{pmatrix}\), then solving the elastic net optimization problem (\ref{obj:primal_elnet}) for a fixed $\tau$, is equivalent to solving the following problem.
\begin{equation}
    \beta(\lambda) \in \argmin_{\beta \in \bbR^p} \frac{1}{2}\norm{\Tilde{y} - \Tilde{X}\beta}_2^2 + \lambda \norm{\beta}_1.
\end{equation}
%
%

%
Now, similar to the LASSO we can write the step-size of inclusion ($\Delta_j^1$) of the $\lambda$-path of ElNet using the augmented data ($\Tilde{X}, \Tilde{y}$) as
\begin{align}\label{eq:step_size_d1_elnet}
    \Delta_j^1 &= \underset{j \in \mathcal{A}_{\lambda_t}^c}{min} \Bigg( \frac{(\Tilde{x}_j - \Tilde{x}_k)^{\top}\Tilde{w}(\lambda_t)}{(\Tilde{x}_j - \Tilde{x}_k)^{\top}\Tilde{v}(\lambda_t)}, \frac{(\Tilde{x}_j + \Tilde{x}_k)^{\top}\Tilde{w}(\lambda_t)}{(\Tilde{x}_j + \Tilde{x}_k)^{\top}\Tilde{v}(\lambda_t)} \Bigg).
\end{align}
However, we cannot just simply augment the data by stacking extra rows as this can be prohibitively expensive due to the combinatorial effects. In order to derive the step-size of inclusion ($\Delta_j^1$) we need a different approach as we construct the high-order interaction model in a progressive manner. We have shown that using the following approach the step-size of inclusion for the $\lambda$-path of ElNet can be computed very efficiently, where the step-size of inclusion can be defined as
\begin{align}
    \Delta_j^1 &= \underset{j \in \mathcal{A}^c_{\lambda_t}}{min} \Bigg( \frac{(x_j - x_k)^{\top}w(\lambda_t) + \alpha \beta_k}{(x_j - x_k)^{\top}v(\lambda_t) - \alpha \nu_k}, \frac{(x_j + x_k)^{\top}w(\lambda_t) - \alpha \beta_k}{(x_j + x_k)^{\top}v(\lambda_t) + \alpha \nu_k} \Bigg).
\end{align}
The derivation of the above step-size ($\Delta_j^1$) is given below.
\begin{proof}
\textnormal{Lets, consider} \(\Tilde{w}(\lambda_t) = \Tilde{y} - \Tilde{X}_{\mcl{A}_{\lambda_t}}\beta_{\mcl{A}_{\lambda_t}}(\lambda_t)  \in \mathbb{R}^{n+p}\) \textnormal{and} \(w(\lambda_t) = y - X_{\mcl{A}_{\lambda_t}}\beta_{\mcl{A}_{\lambda_t}}(\lambda_t)  \in \mathbb{R}^{n}\), \textnormal{where} \(p = |\mathcal{A}_{\lambda_t}| + |\mathcal{A}^c_{\lambda_t}| \), \textnormal{then we can write}
\begin{equation}\label{eqn:aug_w}
    \Tilde{w}_i(\lambda_t) = \begin{cases}
     w_i(\lambda_t) \quad \quad \textnormal{if} \quad  i \leq n, \\
    -\sqrt{\alpha}\beta_j \quad \hspace{0.2cm} \textnormal{if} \quad n < i \leq n + |\mathcal{A}_{\lambda_t}|,\\
    0 \quad \quad \quad \hspace{0.45cm} \textnormal{if} \quad n + |\mathcal{A}_{\lambda_t}| < i \leq n + p.
    \end{cases}
\end{equation}
\textnormal{similarly considering \(\Tilde{v}(\lambda_t) = \Tilde{X}\nu(\lambda_t)  \in \mathbb{R}^{n+p}\) and \(v(\lambda_t) = X\nu(\lambda_t)  \in \mathbb{R}^{n}\), we can write}
\begin{equation}\label{eqn:aug_v}
    \Tilde{v}_i(\lambda_t) = \begin{cases}
     v_i(\lambda_t) \quad \quad \hspace{0.08cm} \textnormal{if} \quad i \leq n ,\\
    \sqrt{\alpha}\nu_j \quad \quad \hspace{0.1cm} \textnormal{if} \quad n < i \leq n + |\mathcal{A}_{\lambda_t}|,\\
    0 \quad \quad \quad \hspace{0.42cm} \textnormal{if} \quad n + |\mathcal{A}_{\lambda_t}| < i \leq n + p.
    \end{cases}
\end{equation}
\textnormal{and, considering \(\Tilde{X} \in \mathbb{R}^{n+p}\) and \(X \in \mathbb{R}^{n}\) we can write}
\begin{equation}\label{eqn:aug_X}
    \Tilde{x}_{ij} = \begin{cases}
    x_{ij} \quad \quad \hspace{0.1cm} \textnormal{if} \quad i \leq n, \\
    \sqrt{\alpha} \quad \quad \hspace{0.03cm} \textnormal{if} \quad i > n \enspace \textnormal{and} \enspace (i-n)=j,\\
    0 \hspace{1.05cm} \textnormal{otherwise}.
    \end{cases}
\end{equation}
\textnormal{Therefore, in (\ref{eq:step_size_d1_elnet}) we can write that \(\forall j \in \mathbb{R}^p\)}
\begin{align*}
    \Tilde{x}_j^{\top} \Tilde{w}(\lambda_t) &= \sum_{i=1}^{n+p} \Tilde{w}_i(\lambda_t) \Tilde{x}_{ij}, \\
    &=\sum_{i=1}^n \Tilde{w}_i(\lambda_t) \Tilde{x}_{ij} + \sum_{\substack{i=n+1}}^{n+|\mathcal{A}_{\lambda_t}|} \Tilde{w}_i(\lambda_t) \Tilde{x}_{ij} + \sum_{i=n+|\mathcal{A}_{\lambda_t}|+1}^{n+p} \Tilde{w}_i(\lambda_t) \Tilde{x}_{ij}.
\end{align*}
\textnormal{Now, using (\ref{eqn:aug_w}) and (\ref{eqn:aug_X}) the second and the third quantity in the above expression can be written as}
\begin{equation*}
\sum_{\substack{i=n+1}}^{n+|\mathcal{A}_{\lambda_t}|} \Tilde{w}_i (\lambda_t) \Tilde{x}_{ij} =\begin{cases}
    (-\sqrt{\alpha}\beta_j)(\sqrt{\alpha}), \hspace{0.5cm} \textnormal{if} \enspace (i-n) = j,\\
    0 \hspace{2.65cm} \textnormal{otherwise}.
    \end{cases}
\end{equation*}
\textnormal{and,}
\begin{equation*}
   \sum_{i=n+|\mathcal{A}_{\lambda_t}|+1}^{n+p} \Tilde{w}_i(\lambda_t) \Tilde{x}_{ij} = 0. 
\end{equation*}
\textnormal{Therefore,}
\begin{equation*}
    \Tilde{x}_j^{\top} \Tilde{w}(\lambda_t) = \sum_{i=1}^n w_i(\lambda_t) x_{ij}, \enspace \forall j \in \mathcal{A}^c_{\lambda_t} \enspace \textnormal{and} \enspace  \Tilde{x}_k^{\top} \Tilde{w}(\lambda_t) = \sum_{i=1}^n w_i(\lambda_t) x_{ik} - \alpha \beta_k, \enspace \forall k \in \mathcal{A}_{\lambda_t}.
\end{equation*}
\textnormal{Similarly, using (\ref{eqn:aug_v}) and (\ref{eqn:aug_X}) we can write}
\begin{equation*}
    \Tilde{x}_j^{\top} \Tilde{v}(\lambda_t)= \sum_{i=1}^n v_i (\lambda_t) x_{ij}, \enspace \forall j \in \mathcal{A}_{\lambda_t}^c \hspace{0.5cm} \textnormal{and} \hspace{0.5cm}  \Tilde{x}_k^{\top} \Tilde{v}(\lambda_t)= \sum_{i=1}^n v_i(\lambda_t) x_{ik} + \alpha \nu_k, \enspace \forall k \in \mathcal{A}_{\lambda_t}.
\end{equation*}
\textnormal{Therefore the step-size of inclusion can be written as}
\begin{align}
    \Delta_j^1 &= \underset{j \in \mathcal{A}_{\lambda_t}^c}{\min} \Bigg( \frac{(x_j - x_k)^{\top}w(\lambda_t) + \alpha \beta_k}{(x_j - x_k)^{\top}v(\lambda_t) - \alpha \nu_k}, \frac{(x_j + x_k)^{\top}w(\lambda_t) - \alpha \beta_k}{(x_j + x_k)^{\top}v(\lambda_t) + \alpha \nu_k} \Bigg).
\end{align}
\end{proof}

\subsubsection{Tree pruning ($\lambda$-path)}
Similar to the LASSO (\ref{eq:lasso_tree_pruning_lmd_path1}) we can use the following inequality in augmented data (\(\Tilde{X}, \Tilde{y}\)) as the pruning criteria for the $\lambda$-path of ElNet.
\begin{equation}\label{cond:pruning}
    \lvert \Tilde{\rho}_{\ell} \lvert + \Delta^{\ast}_{\ell} \lvert \Tilde{\eta}_{\ell} \lvert < \lvert \Tilde{\rho}_k \lvert - \Delta^{\ast}_{\ell} \lvert\Tilde{\eta}_k \lvert,
\end{equation}
where, \(\Tilde{\rho}_{\ell} = \Tilde{x}_{\ell}^{\top}\Tilde{w}(\lambda_t) \) and \(\Tilde{\eta}_{\ell} =\Tilde{x}_{\ell}^{\top} \Tilde{v}(\lambda_t) \quad \forall \ell \in \mathcal{A}^c_{\lambda_t} \), \(\Tilde{\rho}_k = \Tilde{x}_{k}^{\top}\Tilde{w}(\lambda_t) \) and \(\Tilde{\eta}_k = \Tilde{x}_k^{\top} \Tilde{v}(\lambda_t),  \quad \forall k \in \mathcal{A}_{\lambda_t} \) . Now, using (\ref{eqn:aug_w}), (\ref{eqn:aug_v})  and (\ref{eqn:aug_X}) we can show that
\begin{align*}
 \Tilde{\rho}_{\ell}(\lambda_t) = \sum_{i=1}^n w_i(\lambda_t) x_{i\ell}, \enspace  \Tilde{\eta}_{\ell}(\lambda_t) = \sum_{i=1}^n v_i(\lambda_t) x_{i\ell}  \enspace \text{and},\\
 \Tilde{\rho}_k(\lambda_t) = \sum_{i=1}^n w_i(\lambda_t) x_{ik} - \alpha \beta_k,  \enspace \Tilde{\eta}_k(\lambda_t) = \sum_{i=1}^n v_i(\lambda_t) x_{ik} + \alpha \nu_k.
 \end{align*}
Therefore, the pruning condition (\ref{cond:pruning}) can be redefined as -
\begin{multline*}
   |\sum_{i=1}^n w_i(\lambda_t) x_{i\ell}| + \Delta^{\ast}_{\ell} |\sum_{i=1}^n v_i(\lambda_t) x_{i\ell}|  < |\sum_{i=1}^n w_i(\lambda_t) x_{ik}  - \alpha \beta_k| - \Delta^{\ast}_{\ell} |\sum_{i=1}^n v_i(\lambda_t) x_{ik} + \alpha \nu_k|.
\end{multline*}
Now, similar to the LASSO (\ref{eq:lasso_pruning_cond_lmd_path3}) we can also write
\begin{equation}\label{eq:Elnet_pruning_lamda_path}
    b_{w(\lambda_t)} + \Delta^{\ast}_{\ell}  b_{v(\lambda_t)} < |\Bar{\rho}_k(\lambda_t)| - \Delta^{\ast}_{\ell}|\Bar{\eta}_k(\lambda_t)|,
\end{equation}
where,  \(\Bar{\rho}_k(\lambda_t) = \sum_{i=1}^n w_i(\lambda_t) x_{ik} - \alpha \beta_k\), \(\Bar{\eta}_k(\lambda_t) = \sum_{i=1}^n v_i(\lambda_t) x_{ik} + \alpha \nu_k\), and 
\begin{align*}
b_{w(\lambda_t)} &= max \big\{ \sum_{w_i(\lambda_t) < 0} |w_i(\lambda_t)| x_{i\ell}, \sum_{w_i(\lambda_t) > 0} |w_i(\lambda_t)| x_{i\ell} \big\}, \\
b_{v(\lambda_t)} &= max \big\{ \sum_{v_i(\lambda_t) < 0} |v_i(\lambda_t)| x_{i\ell}, \sum_{v_i(\lambda_t) > 0} |v_i(\lambda_t)| x_{i\ell} \big\}.
\end{align*}
Therefore, (\ref{eq:Elnet_pruning_lamda_path}) can be used as the pruning condition for the $\lambda$-path of ElNet.
%
%
%
%
%
\subsection{$\tau$-path: path \wrt to $\tau$ ($\lambda$ fixed)}
If we consider two real values $\tau_t$ and $\tau_{t+1}$ ( $\tau_{t+1}>\tau_t$) at which the active set does not change and their signs also remain the same, then we can write 
\begin{align*}
  \beta_{\mathcal{A}_{\tau_{t}}}(\tau_{t+1}) - \beta_{\mathcal{A}_{\tau_t}}(\tau_t) = \nu_{\mathcal{A}_{\tau_t}}(\tau_t)(\tau_{t+1} - \tau_t),\\ 
   \lambda s_{\mathcal{A}^c_{\tau_{t}}}(\tau_{t+1}) - \lambda s_{\mathcal{A}^c_{\tau_t}}(\tau_t) = \gamma_{\mathcal{A}^c_{\tau_t}}(\tau_t)(\tau_{t+1} - \tau_t).
\end{align*}
where, \(\nu_{\mathcal{A}_{\tau_t}}(\tau_t) = (X_{\mathcal{A}_{\tau_t}}^{\top}X_{\mathcal{A}_{\tau_t}} + \alpha I_{|\mathcal{A}_{\tau_t}|})^{-1}X_{\mathcal{A}_{\tau_t}}^{\top} b \)  and \(\gamma_{\mathcal{A}^c_{\tau_t}}(\tau_t) = X_{\mathcal{A}^c_{\tau_t}}^{\top} b -  X_{\mathcal{A}^c_{\tau_t}}^{\top}X_{\mathcal{A}_{\tau_t}}\nu_{\mathcal{A}_{\tau_t}}(\tau_t)\). 
 Note that here also the only change compared to the LASSO (\ref{LASSO:direction_vector_tau_path}) is the addition of an \(\alpha I_{|\mathcal{A}_{\tau_t}|}\) term to the expression of \(\nu_{\mathcal{A}_{\tau_t}}\). Now, one can also derive a similar expression of step-size of inclusion and deletion as done for the LASSO (\ref{LASSO:step-size_z_path}) by considering the updated expression of \(\nu_{\mathcal{A}_{\tau_t}}(\tau_t)\) and \(\gamma_{\mathcal{A}^c_{\tau_t}}(\tau_t) \).
\subsubsection{Tree pruning ($\tau$-path)}
%
%


Similar to the LASSO (\ref{eq:lemma_3_eq}), by using (\ref{eqn:aug_w}), (\ref{eqn:aug_v})  and (\ref{eqn:aug_X}) the pruning condition for the $\tau$-path of ElNet can be written as
\begin{equation}\label{eqn:elnet_pruning_condn_tau_path3}
    b_{\ell, w(\tau_t)} + \Delta_\ell^\ast b_{\ell, \theta} + \Delta_\ell^\ast b_{\ell, v(\tau_t)} < \lvert \Bar{\rho}_k(\tau_t) \lvert - \Delta_{\ell}^{\ast}\lvert \theta_k \lvert - \Delta_{\ell}^{\ast} \lvert \Bar{\eta}_k(\tau_t) \lvert,
\end{equation}
where  \(\Bar{\rho}_k(\tau_t) = \sum_{i=1}^n w_i(\tau_t) x_{ik} - \alpha \beta_k\), \(\Bar{\eta}_k(\tau_t) = \sum_{i=1}^n v_i(\tau_t) x_{ik} + \alpha \nu_k\).

\section{Additional Results}
\begin{figure}[h!]
   \centering
   \includegraphics[width=0.95\linewidth]{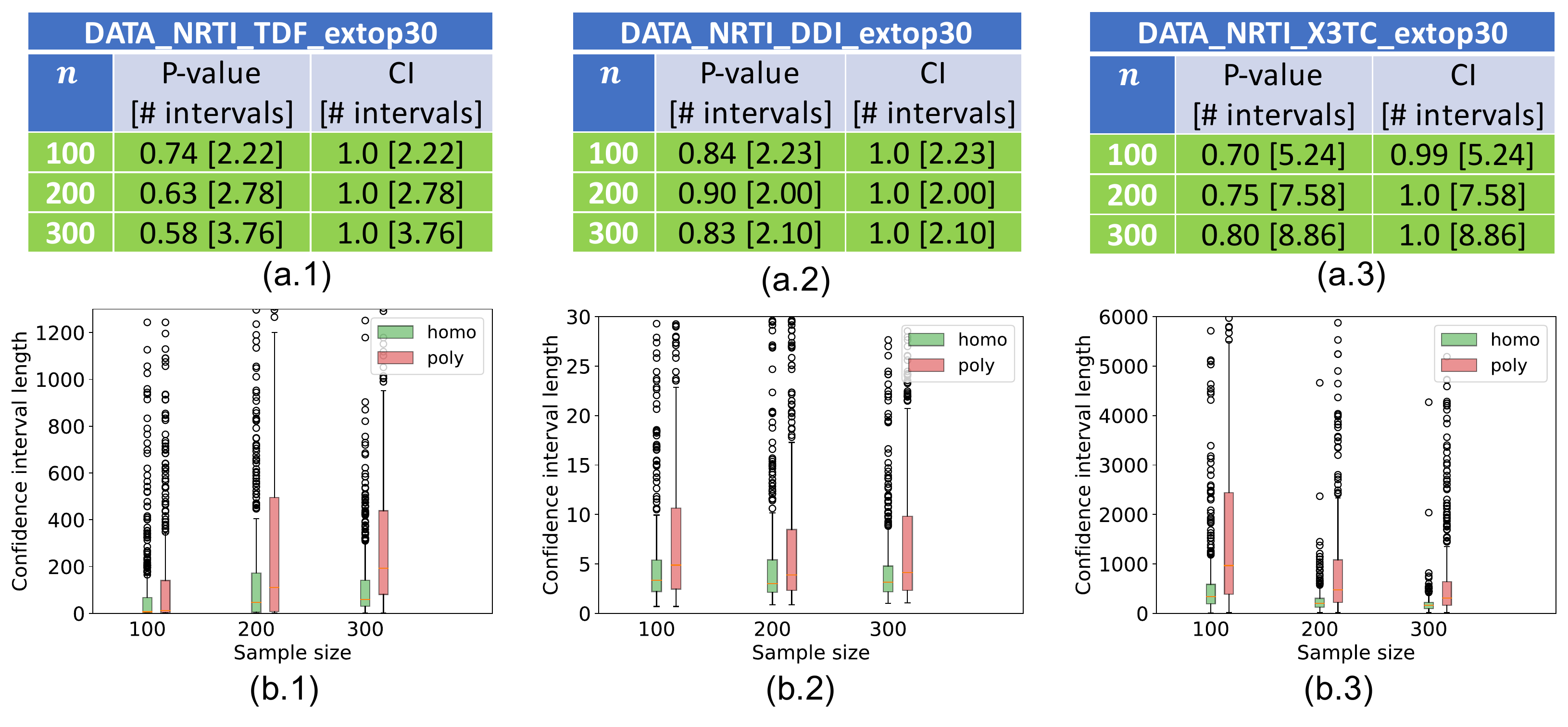}
   \caption{Comparison of statistical powers (Homotopy vs Polytope). (a.1-a.3) show the percentage of cases where selection bias corrected p-values and confidence interval lengths of the proposed method (Homotopy) was smaller than that of the existing method (Polytope) in random sub-sampling experiments. (b.1-b.3) show the distributions of the confidence interval lengths of the same experiments.}
   \label{fig:stats_NRTIs2}
 \end{figure}

\begin{figure}[h!]
   \centering
   \includegraphics[width=0.95\linewidth]{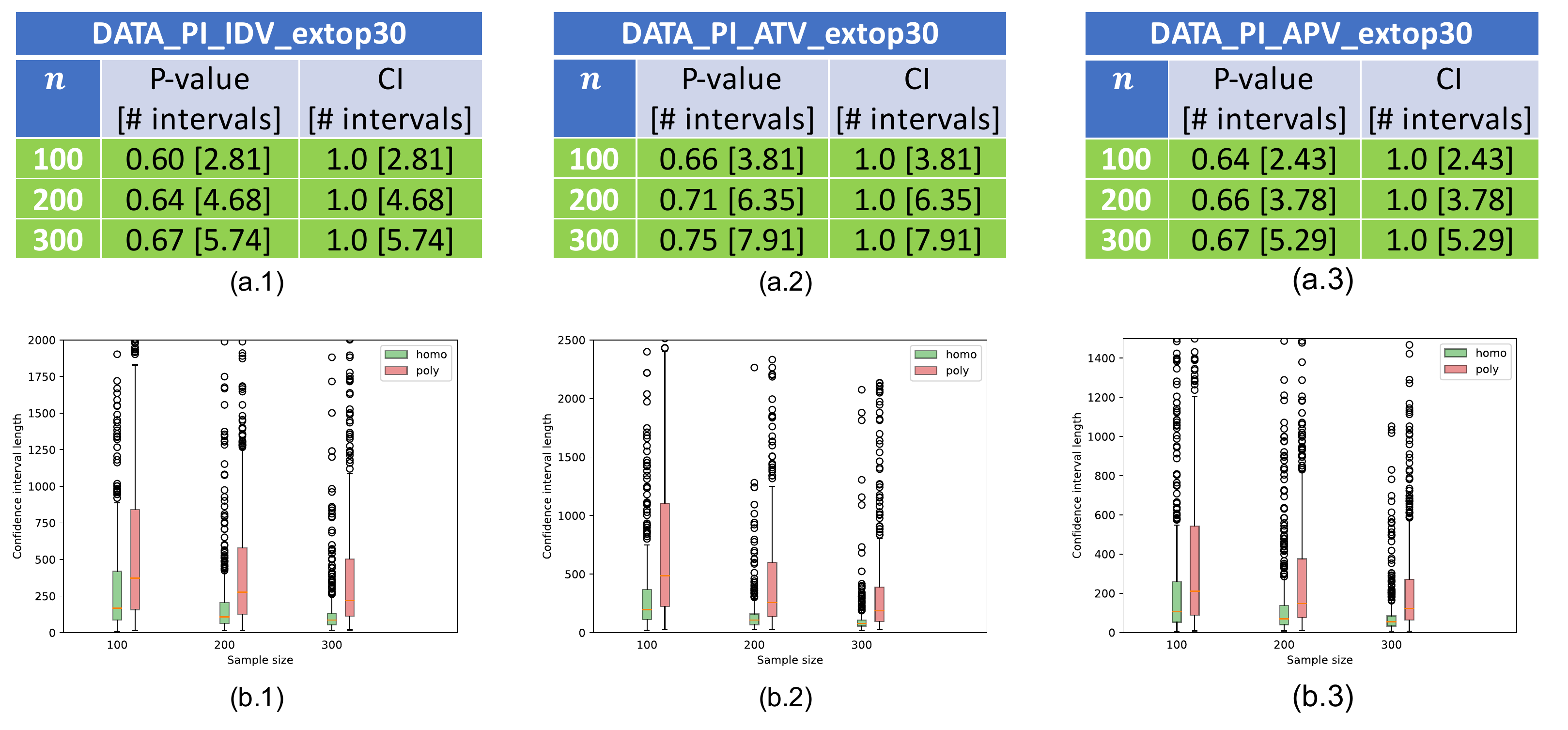}
   \caption{Comparison of statistical powers (Homotopy vs Polytope). (a.1-a.3) show the percentage of cases where selection bias corrected p-values and confidence interval lengths of the proposed method (Homotopy) was smaller than that of the existing method (Polytope) in random sub-sampling experiments. (b.1-b.3) show the distributions of the confidence interval lengths of the same experiments.}
   \label{fig:stats_PIs}
 \end{figure}

\begin{figure}[h!]
   \centering
   \includegraphics[width=0.4\linewidth]{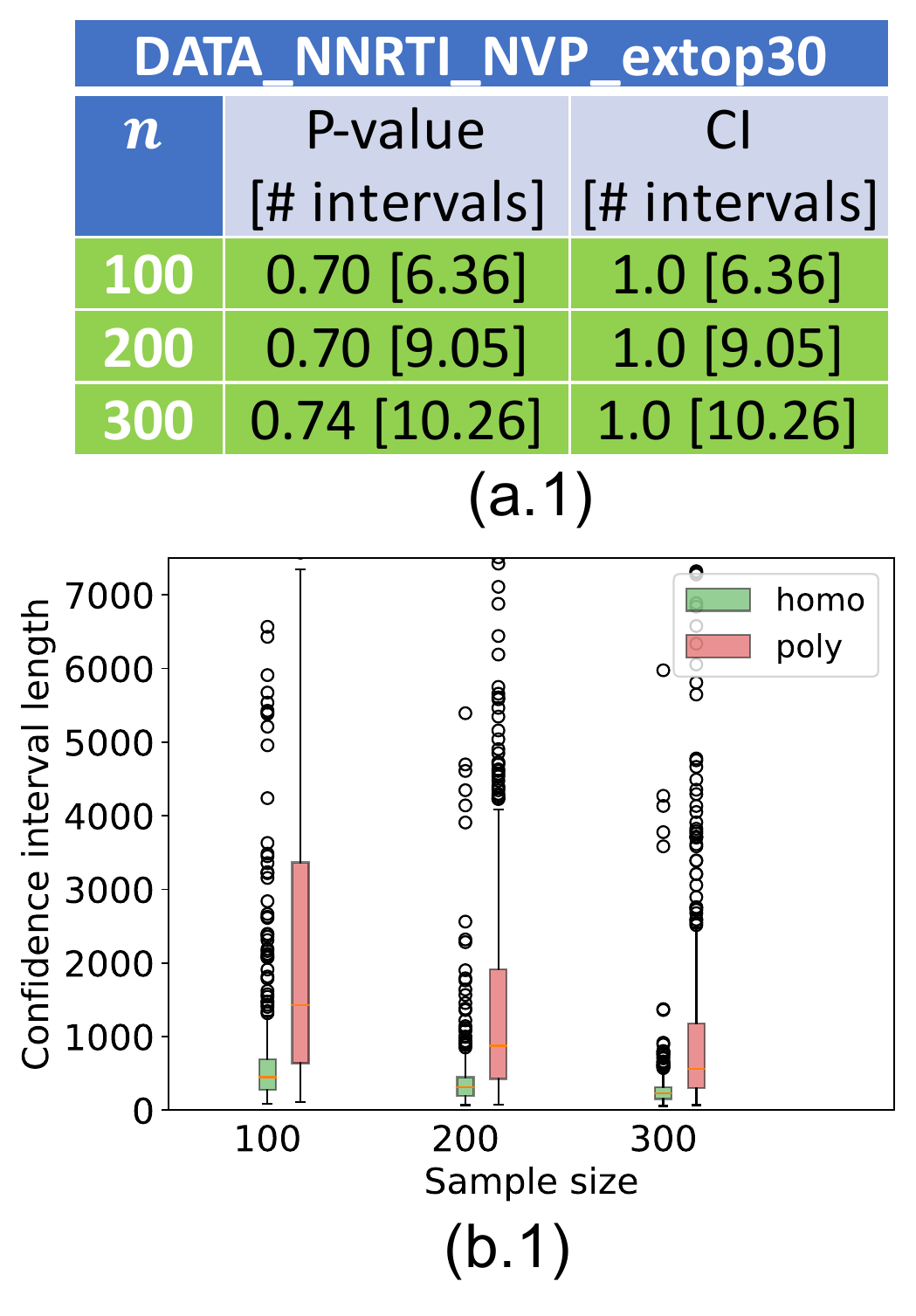}
   \caption{Comparison of statistical powers (Homotopy vs Polytope). (a.1-a.3) show the percentage of cases where selection bias corrected p-values and confidence interval lengths of the proposed method (Homotopy) was smaller than that of the existing method (Polytope) in random sub-sampling experiments. (b.1-b.3) show the distributions of the confidence interval lengths of the same experiments.}
   \label{fig:stats_NNRTI}
 \end{figure}

Here we report additional results using real world HIV-1 sequence data from Stanford HIV Drug Resistance Database \citep{rhee2003human}. This dataset contains three classes of drug data: NRTIs, NNRTIs and PIs consisting of 16 drugs. Finding virus induced mutations which leads to drug resistance is crucial to drug development. However, drug resistance is a complex biological phenomenon and it is often reported in the literature \citep{rhee2006genotypic, ERP_Tsuda, suzumura2017selective}, that it is the association of multiple mutations along with some crucial single mutations that can best describe the phenomenon. Hence, it is important to understand the association of multiple mutations related to the drug resistance. In our experiment we used 6 NRTIs, 1 NNRTIs and  3 PIs drugs. We reported the results on 3 NRTIs drugs in the main article and here we include the results on the remaining 3 NRTIs (Fig. \ref{fig:stats_NRTIs2}) and 3 PIs (Fig. \ref{fig:stats_PIs}) and 1 NNRTI (\ref{fig:stats_NNRTI}) drugs. The continuous drug resistance values corresponds to the response ($y \in \mb{R}$) and the binary mutations corresponds to the original features ($z \in \mb{R}^m$) in our experimental settings.
 In Table.~\ref{table:comp_eff_kink_numbers} we demonstrated the computational advantage of the proposed homotopy method over exiting method on conditioning on model (\cite{lee2016exact}). 
 In this experiment the $\lambda$-path was constructed until the active set ($\mcl{A}$) contains 20 features and subsequently that active set and the corresponding $\lambda$ value is used for the construction of the $\tau$-path.
 The \cite{lee2016exact} method needs to consider the union of all possible signs in the observed active set ($\mcl{A}$) in order to condition on the model. However, our homotopy mining needs to consider only $\sim$ 120 polytopes (worst case) for the same task. 
\begin{table}[h!]
\centering
\begin{tabular}{ |c|c|c| } 
\hline
High-order interactions & \shortstack{Homotopy \\ (\# kinks)} & \shortstack{Polytope \\ (\# polytopes)} \\ 
\hline
$1^{st}$ & $104.15 \pm 10.73$ & $2^{20}$\\ 
\hline
$2^{nd}$ &$101.0 \pm 4.64$  & $2^{20}$\\ 
\hline
$3^{rd}$ &  $78.33 \pm 24.69$ & $2^{20}$\\ 
\hline
\end{tabular}
\caption{Comparison of computational efficiencies of the proposed homotopy method against existing polytope method. The "\# kinks" represents the average number of kinks encountered during the construction of $\tau$-path for each test statistic direction, whereas the "\# polytopes" represents the number of all possible signs one needs to consider to condition on the model.}
\label{table:comp_eff_kink_numbers}
\end{table}

We note that theoretically, in the worst-case, the complexity of the homotopy method grows exponentially. This is a common issue in homotopy-based methods such as computing regularization paths. However, fortunately, it has been well-recognized \citep{le2021parametric} that this worst case rarely happens in practice, and this is also evident from our experimental results.

Similar to the pruning, empirical evidence also demonstrates that homotopy is more efficient in case of high-order interaction terms compared to that of singleton terms, and the efficiency increases as the order of interaction increases. 
We suspect that as the order of interaction increases the sparsity of the data also increases which significantly affects the construction of the $\tau$-path as evident from the effectiveness of both pruning and the homotopy method. However, more theoretical investigations are required to have a clear understanding of this phenomenon which we believe worth considering in the future.


\end{document}